%% file: devon.tex
\documentclass[10pt,twocolumn,letterpaper]{article}

\usepackage{bm}
\usepackage{url}
\usepackage{float} 
\usepackage{listings}
\usepackage{subfig}
\usepackage{cvpr}
\usepackage{times}
\usepackage{epsfig}
\usepackage{graphicx}
\usepackage{amsmath}
\usepackage{amssymb}
\usepackage{setspace}
\usepackage{lipsum}
\usepackage{color}
\usepackage{multirow}
\usepackage{tikz}
\usetikzlibrary{arrows}
\tikzstyle{int}=[draw, fill=blue!20, minimum size=2em]
\tikzstyle{init} = [pin edge={to-,thin,black}]
\usetikzlibrary{shapes.geometric}
\usetikzlibrary{decorations.text}
\usetikzlibrary{calc}
\usetikzlibrary{arrows.meta}
\usetikzlibrary{positioning}

\tikzset{module/.style={fill=white!10,auto,align=center,thick,draw,minimum width=0.4cm,minimum height=0.4cm}}

\tikzset{op/.style={fill=black,draw,circle,inner sep=0.0cm,minimum width=0.1cm,minimum height=0.1cm}}

\cvprfinalcopy

\def\cvprPaperID{1906} 

\ifcvprfinal\fi

\begin{document}

\title{Devon: Deformable Volume Network for Learning Optical Flow}

\author{
Yao Lu\\
Data61, ANU
\and
Jack Valmadre\\
University of Oxford
\and
Heng Wang\\
Facebook
\and
Juho Kannala\\
Aalto University
\and
Mehrtash Harandi\\
Data61, Monash University
\and
Philip H. S. Torr\\
University of Oxford
}

\maketitle

\begin{abstract}
State-of-the-art neural network models estimate large displacement optical flow in multi-resolution and use warping to propagate the estimation between two resolutions. Despite their impressive results, it is known that there are two problems with the approach. First, the multi-resolution estimation of optical flow fails in situations where small objects move fast. Second, warping creates artifacts when occlusion or dis-occlusion happens. In this paper, we propose a new neural network module, Deformable Cost Volume, which alleviates the two problems. Based on this module, we designed the Deformable Volume Network (Devon) which can estimate multi-scale optical flow in a single high resolution. Experiments show Devon is more suitable in handling small objects moving fast and achieves comparable results to the state-of-the-art methods in public benchmarks.
\end{abstract}

\section{Introduction}

Optical flow estimation is the problem of finding pixel-wise motions between consecutive images. It is a classic problem in computer vision  and has been studied for more than 30 years. Since Horn and Schunck's variational method~\cite{horn1981determining}, a large number of optical flow algorithms~\cite{beauchemin1995computation,fleet2006optical,fortun2015optical} has been developed. Nevertheless, the problem of estimating optical flow is not yet solved. One can even claim that we still do not have a practical algorithm which is fast, robust and accurate for real-world images.

Recently, supervised learning of optical flow with neural networks has shown great promises~\cite{fischer2015flownet,ilg2016flownet,ranjan2016optical,teney2016learning,thewlis2016fully,sun2017pwc,hui2018liteflownet}. 
By making use of graphics engines, large scale datasets with synthetic images and ground-truth optical flow can be  generated~\cite{butler2012naturalistic,fischer2015flownet,mayer2016large}, which in return enable us to learn optical flow in an end-to-end manner.
We note that while the training images are synthetic, neural networks seem to generalize reasonably well on 
real-world images ~\cite{mayer2018makes}. Compared to classical methods, the neural network approaches have the advantage of offering robust features and fast inference~\cite{fischer2015flownet,ilg2016flownet,sun2017pwc,hui2018liteflownet}.

One of the major difficulties in estimating optical flow is large displacements. 
Learning optical flow with large displacements cannot be achieved by Convolutional Neural Networks (CNNs) with only standard convolution operations. This is due to the fact that CNNs make use of small filters (sizes are rarely larger than $11\times 11$). One cannot afford to increase filter sizes na\"ively to cover large displacements, as the number of parameters and the computational cost both increase drastically.

To handle large displacements, multi-resolution estimation of optical flow is employed in several neural network models. SpyNet \cite{ranjan2016optical} downsamples the original images into multiple resolution levels and each level is handled by a CNN decoder to output optical flow of the corresponding resolution. 
PWC-Net \cite{sun2017pwc} and LiteFlowNet \cite{hui2018liteflownet} follow the same strategy except that they use strided convolutions instead of downsampling to reduce the resolution and use cost volumes as a more explicit representation of motion. As a result, a decoder in lower resolution has effectively a larger receptive size to cover large displacements. However, as pointed out in \cite{yang2017s2f}, the multi-resolution estimation of optical flow faces the ``small objects move fast" problem as small objects disappear in lower resolution and cannot be recovered in higher resolution due to their large motion \cite{brox2004high,sevilla2014optical}. Although the problem is not severe in terms of performance on the current public benchmarks, it limits the use of optical flow for high precision applications where small objects might contain vital information. The key to solve the problem is to handle multi-scale motion in a single high resolution~\cite{yang2017s2f}.

Another technique to handle large displacements in neural network models such as~\cite{ilg2016flownet,ranjan2016optical,sun2017pwc} is warping, which propagates optical flow between two stages in a model. However, warping often creates distortions and artifacts. This issue will be discussed in details in \textsection\ref{sec:problem_warping}.

In this paper, we introduce the Deformable Volume Network (Devon), which avoids the drawbacks of multi-resolution estimation and warping.

\section{The Problem of Warping}
\label{sec:problem_warping}

Warping has been used in variational methods \cite{brox2004high,lucas1981iterative} and  neural network models \cite{ilg2016flownet,ranjan2016optical,sun2017pwc} for iteratively refining optical flow estimations in a multi-stage framework.
The first stage covers large displacements and outputs a rough estimation. Then the second image (or its feature maps) is warped by the roughly estimated optical flow such that pixels of large displacements in the second image are moved closer to their correspondences in the first image. As a result, the next stage, which receives the original first image and the warped second image as inputs, only needs to handle smaller
displacements to refine the estimation.

Let  $I: \mathbb{R}^2 \rightarrow \mathbb{R}^3$ 
denote the first image, $J: \mathbb{R}^2 \rightarrow \mathbb{R}^3$ denote the second image and $F: \mathbb{R}^2 \rightarrow \mathbb{R}^2$ denote the optical flow field of the first image. The warped second image is defined as 
\begin{align}
\tilde{J}(\mathbf{p}) = J(\mathbf{p} + F(\mathbf{p}))
\label{eq:warp}
\end{align}
for image location $\mathbf{p} \in \mathbb{R}^2$~\cite{ilg2016flownet}.

\begin{figure}[t!]
\centering
\subfloat[First image]{
\includegraphics[width=0.4\columnwidth]{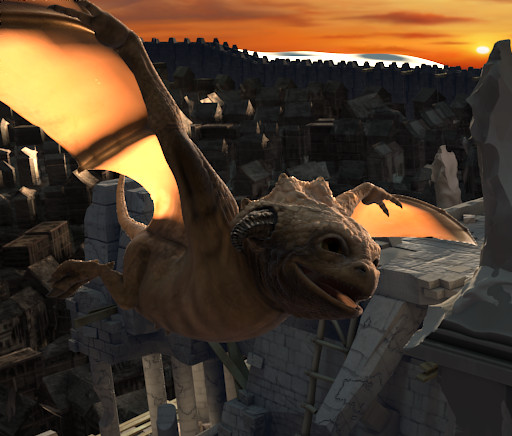}}
\vspace{0.2cm}
\subfloat[Second image]{
\includegraphics[width=0.4\columnwidth]{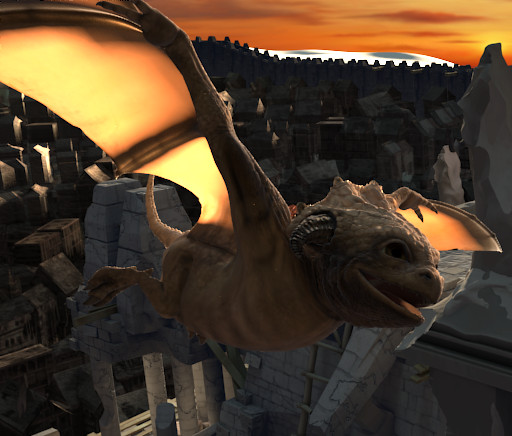}}

\subfloat[Ground truth optical flow]{
\includegraphics[width=0.4\columnwidth]{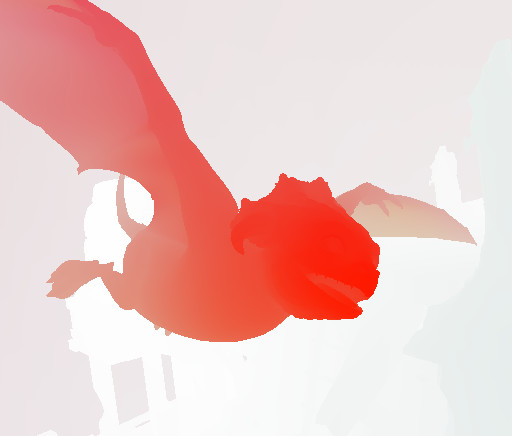}}
\vspace{0.2cm}
\subfloat[Warped second image]{
\includegraphics[width=0.4\columnwidth]{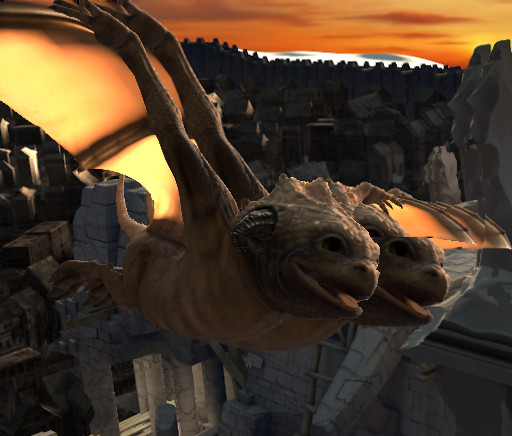}}

\subfloat[Warped second image subtracted by the first image]{
	\includegraphics[width=0.4\columnwidth]{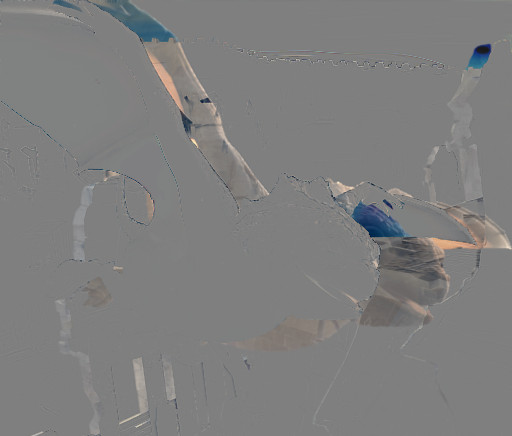}}
\vspace{0.2cm}
\subfloat[Warped second image subtracted by the second image]{
	\includegraphics[width=0.4\columnwidth]{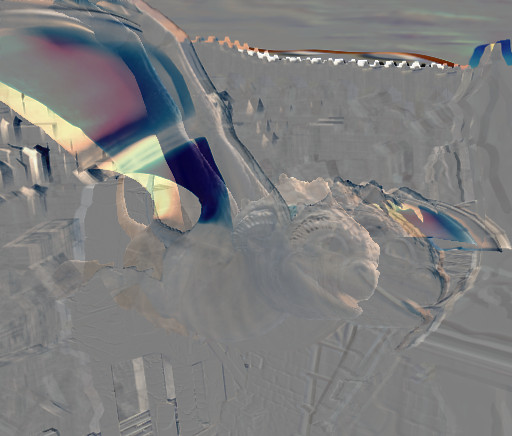}}
\caption{Artifacts of using image warping. From (d), we can see the duplicates of the dragon head and wings. The images and the ground truth optical flow are from the Sintel dataset \cite{butler2012naturalistic}. Warping is done with function \texttt{image.warp()} in the Torch-image toolbox.}
\label{warping}
\end{figure}

The warping operation creates a transformed image reasonably well if the new pixel locations $\mathbf{p}+F(\mathbf{p})$ do not occlude or collide with each other. For example, this is the case with the affine transform $F(\mathbf{p}) = \mathbf{Ap} + \mathbf{t}$ where $\mathbf{A}$ and $\mathbf{t}$ are the transformation parameters. 
However, for real-world images, occlusions are common (\eg, when an object moves and the background is still). If an image is warped with the optical flow which induces occlusions, duplicates will be created. 
The effect is demonstrated in Figure~\ref{warping}. The artifacts cannot be cleaned simply by subtracting the first or the second image from the warped image, as shown in Figure~\ref{warping} (e) and (f).  The artifacts induced by warping have been previously observed 
in~\cite{brox2009large,wang2018occlusion,janai2018unsupervised}.
Intuitively, if a pixel which is moved by warping to a new location and no other pixel are moved to fill in its old location, the pixel will appear twice in the warped image.

Mathematically, consider the following example. Assume the value of $J(\mathbf{p}_1)$ is unique in $J$, that is, $J(\mathbf{p}) \neq  J(\mathbf{p}_1)$ for all $\mathbf{p} \neq \mathbf{p}_1$. 
Then for an optical flow field in which
\begin{align}
F(\mathbf{p}_1) = 0, \quad  F(\mathbf{p}_2) = \mathbf{p}_1-\mathbf{p}_2,
\end{align}
we have
\begin{align}
\tilde{J}(\mathbf{p}_1) &= J(\mathbf{p}_1 + F(\mathbf{p}_1)) \\
 &= J(\mathbf{p}_1 + 0)  =  J(\mathbf{p}_1),  \\
\tilde{J}(\mathbf{p}_2) &= J(\mathbf{p}_2 + F(\mathbf{p}_2)) \\
&= J(\mathbf{p}_2 + \mathbf{p}_1 - \mathbf{p}_2) =  J(\mathbf{p}_1).
\end{align}
Therefore $\tilde{J}(\mathbf{p}_1) = \tilde{J}(\mathbf{p}_2) = J(\mathbf{p}_1)$. Since the value of $J(\mathbf{p}_1)$ is unique in image $J$ but not unique in $\tilde{J}$, a duplicate is created on the warped second image $\tilde{J}$.

When the duplicates happen, it makes the optical flow estimation erroneous since artificial candidate correspondences are created.
Although neural networks as universal approximators might be able to learn the self-corrected correspondences if trained with ground-truth optical flow, one should be aware of the issue which might cause problems in designing non-learning-based methods or more interpretable neural network models.

\section{Deformable Cost Volume}

Let $I$ denote the first image, $J$ denote the second image and $f_I: \mathbb{R}^2 \rightarrow \mathbb{R}^d$ and $f_J: \mathbb{R}^2 \rightarrow \mathbb{R}^d$ denote their feature maps of dimensionality $d$, respectively. The standard cost volume is defined as
\begin{align}
C(\mathbf{p},\mathbf{v}) = \|f_I(\mathbf{p}) -f_J(\mathbf{p}+\mathbf{v}) \|, \label{cv}
\end{align}
for image location $\mathbf{p} \in \mathbb{R}^2$, neighbor $\mathbf{v} \in [-\frac{k-1}{2},\frac{k-1}{2}]^2$ of neighborhood size $k$ and a given vector norm $\|\cdot \|$.

The cost volume gives an explicit representation of displacements. The idea of using cost volume goes back to stereo matching \cite{scharstein2002taxonomy}. When using the feature maps learned by neural networks, construction and processing of a fully connected cost volume, in which the neighborhood is large enough to cover the maximum displacement, leads to high performance in stereo matching \cite{zbontar2016stereo} and optical flow  \cite{XuCVPR2017DCFlow}. 
However, the displacements in stereo matching are one-dimensional while in optical flow they are two-dimensional.
For two images (and their feature maps) of resolution $m\times n$, 
the construction of the cost volume in equation (\ref{cv}) has time and space complexity $O(mndk^2)$. Na\"ively increasing neighborhood size $k$ to cover large displacements increases the computation quadratically. As a result,
DCFlow requires several seconds to compute optical flow for a pair of images on a GPU and large memory usage \cite{XuCVPR2017DCFlow}. 

To reduce the computational burden, in  \cite{sun2017pwc,hui2018liteflownet},  multi-resolution feature maps of two images are created and a cost volume of a small neighborhood is constructed at each resolution. Warping is used to propagate the optical flow between two resolutions.
However, as discussed before, the multi-resolution estimation of optical flow leads to the ``small objects move fast" problem and warping induces artifacts and distortion.
To avoid the drawbacks, we propose a new neural network module, the deformable cost volume. The key idea is:
instead of deforming images or their feature maps, we deform the cost volume and leave the images and the feature maps unchanged.

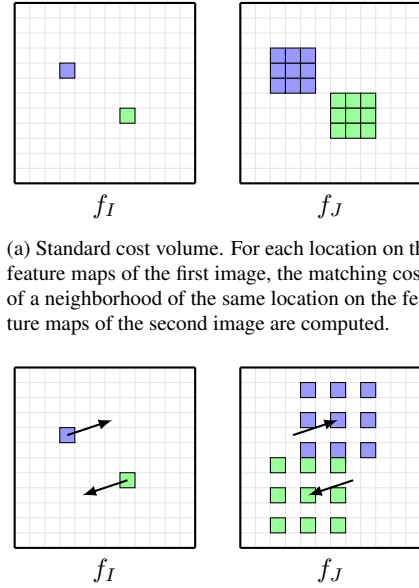
\begin{figure}[t!]
\centering

\subfloat[Standard cost volume. For each location on the feature maps of the first image, the matching costs of a neighborhood of the same location on the feature maps of the second image are computed.]{
\input{img/scv.tex}
}

\subfloat[Deformable cost volume. For each location on the feature maps of the first image, the matching costs of a \textbf{dilated} neighborhood of the same location, \textbf{offset by a flow vector}, on the feature maps of the second image are computed.]{
\input{img/dcv.tex}
}	

\vspace{0.2cm}	
\caption{Cost Volumes}	
\label{dcv}
\end{figure}

The proposed deformable cost volume is defined as 
\begin{align}
C(\mathbf{p},\mathbf{v},r,F) 
= \|f_I(\mathbf{p}) -f_J(\mathbf{p}+r\cdot\mathbf{v} + F(\mathbf{p})) \|
\label{eq:dcv}
\end{align}
where $r$ is the dilation rate and $F(\cdot)$ is an external flow field. The dilation rate $r$ is introduced to enlarge the size of the neighborhood to handle large displacements without reducing the resolution. This is inspired by the dilated convolution \cite{chen2016deeplab,yu2015multi} which enlarges its receptive field in a similar way. $F(\cdot)$ can be obtained from the optical flow estimated from a previous stage or an external algorithm. If $F(\mathbf{p})=0$ for all $\mathbf{p}$ and $r=1$, then the deformable cost volume is reduced to the standard cost volume. 
For non-integer $F(\mathbf{p})$, bilinear interpolation is used.  The deformable cost volume is illustrated in Figure \ref{dcv}.

Since the deformable cost volume does not distort $f_I$ or $f_J$, the artifacts associated with warping will not be created. Optical flow can be inferred from the deformable cost volume solely without resorting to the feature maps of the first image to counter the duplicates.

The deformable cost volume is differentiable with respect to $f_I(\mathbf{p})$ and $f_J(\mathbf{p}+r\cdot\mathbf{v} + F(\mathbf{p}))$ for each image location $\mathbf{p}$. Due to bilinear interpolation, the deformable cost volume is also differentiable with respect to $F(\mathbf{p})$, using the same technique as in \cite{ilg2016flownet,jaderberg2015spatial}. Therefore, the deformable cost volume can be inserted in a neural network for end-to-end learning optical flow.

To see how the deformable cost volume avoids the artifacts of warping, consider the following. Assume  $F(\mathbf{p}_1) = \mathbf{0}$ and $F(\mathbf{p}_2) = \mathbf{p}_1-\mathbf{p}_2$.
The standard cost volume (\ref{cv}) with warping (\ref{eq:warp}) gives
\begin{align}
C(\mathbf{p}_1,\mathbf{0}) &= \|f_I(\mathbf{p}_1) -\tilde{f_J}(\mathbf{p}_1+ \mathbf{0}) \| \\
&= \|f_I(\mathbf{p}_1) -f_J(\mathbf{p}_1+F(\mathbf{p}_1)) \|\\
&= \|f_I(\mathbf{p}_1) -f_J(\mathbf{p}_1) \| \\
C(\mathbf{p}_1,\mathbf{p}_2-\mathbf{p}_1) &= \|f_I(\mathbf{p}_1) -\tilde{f_J}(\mathbf{p}_1+\mathbf{p}_2-\mathbf{p}_1) \| \\
&= \|f_I(\mathbf{p}_1) -\tilde{f_J}(\mathbf{p}_2) \| \\
&= \|f_I(\mathbf{p}_1) -f_J(\mathbf{p}_2+F(\mathbf{p}_2)) \|\\
&= \|f_I(\mathbf{p}_1) -f_J(\mathbf{p}_1) \| 
\end{align}
That $C(\mathbf{p}_1,\mathbf{0}) = C(\mathbf{p}_1,\mathbf{p}_2-\mathbf{p}_1)$ implies $f_I(\mathbf{p}_1)$ has the same matching cost for $f_J(\mathbf{p}_1)$ and $f_J(\mathbf{p}_2)$, which does not hold in general and makes the matching ambiguous. 
On the other hand, with deformable cost volume (\ref{eq:dcv}) of dilation rate one, we have
\begin{align}
C(\mathbf{p}_1,\mathbf{0}) &= \|f_I(\mathbf{p}_1) -f_J(\mathbf{p}_1) \| \\
C(\mathbf{p}_1,\mathbf{p}_2-\mathbf{p}_1) &= \|f_I(\mathbf{p}_1) -f_J(\mathbf{p}_2+F(\mathbf{p}_1)) \|\\
&= \|f_I(\mathbf{p}_1) -f_J(\mathbf{p}_2) \| 
\end{align}
As $C(\mathbf{p}_1,\mathbf{0}) \neq C(\mathbf{p}_1,\mathbf{p}_2-\mathbf{p}_1)$ in general, the artifact is avoided.

\newpage

\section{Deformable Volume Network}\label{sec:devon}

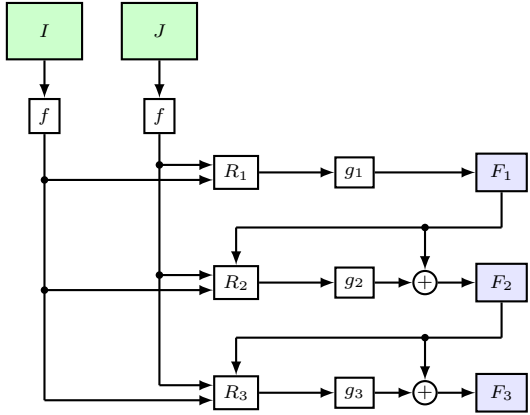
\begin{figure}[t!]
\centering
\input{img/devon_archi.tex}
\vspace{0.2cm}
\caption{Deformable Volume Network (Devon) with three stages. $I$ denotes the first image, $J$ denotes the second image, $f$ denotes the encoding module (\textsection\ref{sec:f}), $R_t$ denotes the relation module (\textsection\ref{sec:R}),  $g_t$ denotes the decoding module (\textsection\ref{sec:g})  and $F_t$ denotes the estimated optical flow for stage $t$.}
\label{devon}
\end{figure}

Our proposed model is the Deformable Volume Network (Devon), as illustrated in Figure \ref{devon}. Devon has multiple stages. Each stage is  a neural network with an identical Siamese architecture, which  consists of an encoding module, a relation module and a decoding module.  Each stage outputs the optical flow field of 1/4 resolution and then the flow from last stage is bilinearly upsampled to obtain the final prediction.
The optical flow estimated from a previous stage is propagated to the current one through the deformable cost volume and residual connections.

Compared to previous neural network models~\cite{fischer2015flownet,ilg2016flownet,ranjan2016optical,sun2017pwc,hui2018liteflownet}, Devon is different in the following major ways: (1) All stages in Devon output optical flow of the same resolution. While Devon handles multi-scale motion by the deformable cost volume, it does not use the multi-resolution representation of motion. Extensive downsampling leads to the loss of information and deficiency in handling the ``small objects move fast" problem. The advantage of estimating motion direct on a single high resolution is also shown in \cite{yang2017s2f}. 
(2) Each stage acts on the undistorted images. No warping is used. Therefore, the artifacts discussed in  \textsection\ref{sec:problem_warping} can be avoided. (3) The decoding module only receives inputs
from the relation module. Therefore, neural networks
infer the optical flow solely from the relations between two
images, rather than memorize the optical flow pattern of
a single image as a short-cut. The short-cut issue has appeared
when applying neural networks to learn monocular
stereo \cite{ummenhofer2016demon}. On the contrary, in FlowNetC, PWC-Net and LiteFlowNet, 
the decoding module also receives inputs from the encoding module of the first image. (4) The encoding module is shared in all stages.

\newpage

We describe the details of each module structure below. All convolution layers have zero-padding size one. Besides, all convolution layers, except the last one in the encoding module and the last one in the decoding module, are followed by a leaky ReLU function~\cite{he2015delving} with leakiness 0.1.

\subsection{Encoding Module}\label{sec:f}
\begin{figure}[h!]	
\centering
\input{img/f_rich.tex}
\caption{Encoding module $f$. The residual connection denotes the output of a layer is added to the output of another layer.}		
\label{f}
\end{figure}
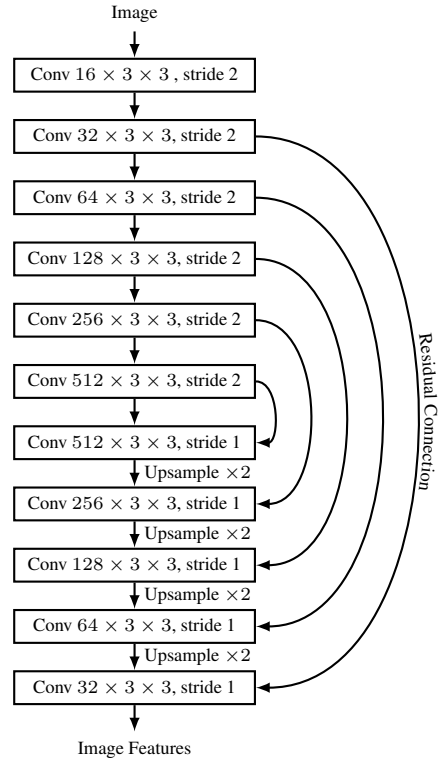

The encoding module has a U-Net structure \cite{ronneberger2015u} with residual connections \cite{he2016deep}, as shown in Figure \ref{f}. Since the model output optical flow of 1/4 resolution, the encoding module has 6 convolution layers of stride 2 but 4 upsampling layers. We also experimented with a simpler module and result is shown in the ablation analysis in \textsection\ref{sec:ablation}.

\subsection{Relation Module}\label{sec:R}

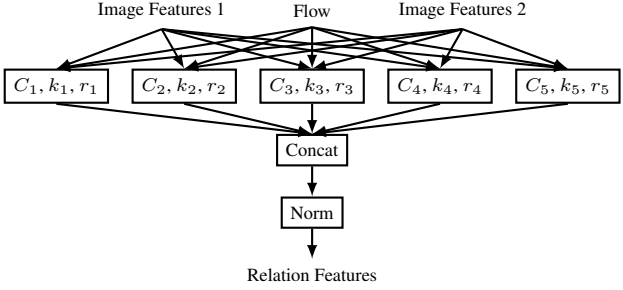
\begin{figure}[h!]	
\centering	
\input{img/R.tex}
\caption{Relation module $R$. $C_1\sim C_5$ denote the deformable cost volumes. $k_1\sim k_5$  denote the neighborhood sizes. $r_1\sim 
r_5$  denote the dilation rates. Concat denotes concatenation. Norm denotes normalization.}	
\label{fig:C}
\end{figure}

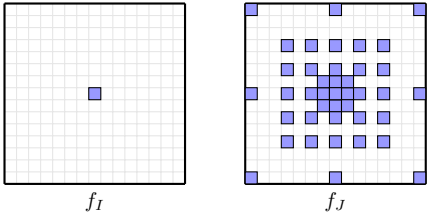
\begin{figure}[h!]
    \centering
    \input{img/retina.tex}
    \caption{Concatenation of deformable cost volumes creates a retinal structure of correspondences. In this example, three cost volumes of 
neighborhood sizes $(k_1,k_2,k_3)=(3,5,3)$ and dilation rates $(r_1,r_2,r_3)=(1,2,7)$ respectively are concatenated.}
    \label{fig:retina}
    \vspace{0.2cm}
\end{figure}

The relation module is illustrated in Figure \ref{fig:C}. It concatenates the outputs of five deformable cost volumes, which have different neighborhood size $k$ or dilation rate $r$, as shown in Table \ref{table:dcv}. Such combination enables dense correspondences nearby the center of an image location and sparse correspondences in peripheral to capture multi-scale motion. This is consistent with the fact that small displacements are more frequent in natural videos \cite{roth2007spatial} and resembles the structure of retina, as illustrated in Figure \ref{fig:retina}.

\newpage

\begin{table}[h!]
	\begin{center}
    \begin{small}
		\begin{tabular}{|c|c|c|}
			\hline
			    & $(k_1,k_2,k_3,k_4,k_5)$  & $(r_1,r_2,r_3,r_4,r_5)$   \\
			\hline
    			$R_1$ & $(5,5,5,5,9)$ &  $(1,3,8,12,20)$ \\
    			$R_2$ & $(5,5,5,5,9)$ &  $(1,3,8,10,12)$ \\
    			$R_3$ & $(5,5,5,5,9)$ &  $(1,3,4,5,7)$ \\
			\hline
		\end{tabular}    
    \end{small}
	\end{center}
	\caption{Hyperparameters of deformable cost volumes in Devon.}
    \label{table:dcv}
\end{table}

Since Devon is a multi-stage model which performs coarse-to-fine estimation of optical flow, we gradually decrease the dilation rates as the optical flow is expected to get finer in later stages.  We also experimented with using the cost volumes of the same neighbor sizes and dilation rates for all three stages, the result of which is reported in \textsection\ref{sec:ablation}.

Next, for each location in the concatenated feature maps, a normalization method is applied across the channels. We apply $f(C)=\exp(-C)$ elements-wise, which encourages the elements representing the minimum cost to pop-up. Such normalization improves the estimation accuracy as shown in \textsection\ref{sec:ablation}.

The output of this module has size $(k_1^2+k_2^2+k_3^2+k_4^2+k_5^2)\times m\times n$, where $m$ is the height and $n$ is the width of the feature maps. When the module does not receive an optical flow as one
of the inputs (in the first stage), it is set to receive a zero-valued optical flow field. 

\newpage

Since the relation module captures multi-scale motion without reducing the resolution, the presence and precise location of small objects which move fast are retained. This allows Devon to have better chance in solving the ``small objects move fast" problem.

\subsection{Decoding Module}\label{sec:g}
\begin{figure}[h!]
\centering
\input{img/g.tex}
\vspace{0.2cm}
\caption{Decoding module $g$. The residual connection denotes the output of a layer is added to the output of another layer.}
\label{fig:g}
\end{figure}
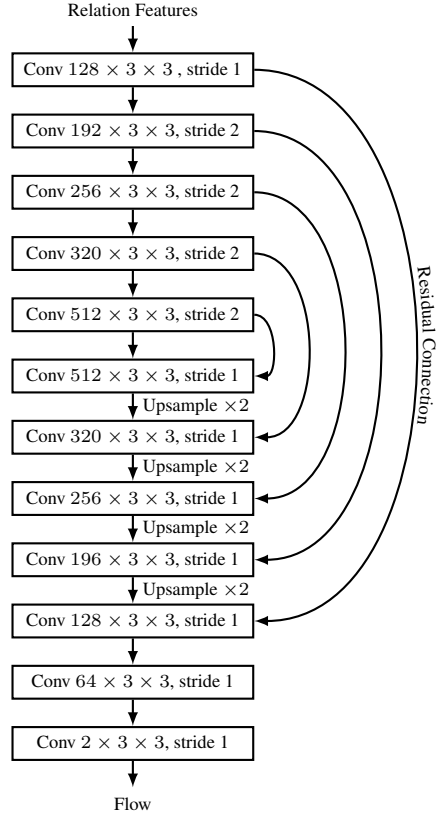

In the decoding module, we again use the U-Net structure with residual connections. The whole module structure is illustrated in Figure \ref{fig:g}.  Each stage has its own decoder. We also  experimented with sharing decoder in all three stages. The result is reported in \textsection\ref{sec:ablation}.

\section{Experiments}
We evaluate Devon on two challenging benchmarks which contain large motions: Sintel \cite{butler2012naturalistic} and KITTI \cite{Geiger2012CVPR}, as in \cite{fischer2015flownet, ranjan2016optical,ilg2016flownet,sun2017pwc}.
We compare Devon with the previous neural network models: FlowNetS \cite{fischer2015flownet}, FlowNetC \cite{fischer2015flownet}, FlowNet2 \cite{ilg2016flownet}, SpyNet \cite{ranjan2016optical}, PWC-Net \cite{sun2017pwc} and LiteFlowNet \cite{hui2018liteflownet}.
We use Devon with three stages. We use $l_1$ norm for the deformable cost volumes.

\subsection{Training}\label{sec:training}
Our training procedure largely follows from \cite{ilg2016flownet,sun2017pwc}. We first train our network on FlyingChairs. We use the $l_2$ loss function
\begin{align}
L = \sum_t \gamma_t | F_{\text{GT}} - \hat{F_t}  |_2
\end{align}
where $F_{\text{GT}}$ denotes the ground-truth optical flow and $\hat{F_t}$  denotes the bilinear upsampled network output at stage $t$. For Devon with three stages, we choose $\gamma_1 = 0.2$, $\gamma_2 = 0.3$ and $\gamma_3 = 0.5$.  All pixel values of the images are multiplied by $1/255$. Empirically, such normalization is found to accelerate the training.
For optimization algorithm, we use Adam \cite{kingma2014adam} with $\beta_1 = 0.9$, $\beta_2 = 0.999$ and weight decay factor 0.0004.
We use the weight initialization method in \cite{he2015delving}.
We use data augmentation which is consisted of random cropping of size 448 $\times$ 384, translation, rotation, color jittering and Gaussian noise. There are totally 500000 mini-batch updates with mini-batch size 8. The learning rate starts from $1e-4$ and halves at 200000, 300000 and 400000  mini-batch updates. 

After  training on FlyingChairs, we fine-tune our model on FlyingThings3D  (final pass) with the robust loss function
\begin{align}
L = \sum_t \gamma_t ( |F_{\text{GT}} - \hat{F_t} |_1 + \epsilon )^q
\end{align}
where $q=0.4$ and $\epsilon=0.01$. There are totally 500000 mini-batch updates with mini-batch size 4. We use data augmentation which is consisted of random cropping of size 786 $\times$ 384, translation and color jittering. We do not use Gaussian noise during data augmentation since the images have motion blur already. The learning rate starts from $1e-5$ and halves at 200000, 300000 and 400000 mini-batch updates. 

For the evaluation on Sintel, we fine-tune the model, which is trained on FlyingChairs and fine-tuned on FlyingThings3D,  on the Sintel training set. We use the robust loss function described above and learning rate schedule in~\cite{sun2018models}. The data augmentation consists of random cropping of size 786 $\times$ 384, rotation, horizontal flipping and color jittering. The mini-batch size is 4.

For the evaluation on KITTI, we fine-tune the network initially fine-tuned on FlyingThings3D, on the KITTI training set. We use  the robust loss function described above and learning rate schedule in~\cite{sun2018models}. The data augmentation consists of random cropping of size  896 $\times$ 320, horizontal flipping and color jittering. The mini-batch size is 4. We use the same learning schedule as the one for Sintel. The training set is mixed with KITTI 2012 and KITTI 2015. Invalid pixels are excluded in computing the loss.

All the experiments are conducted with PyTorch. The deformable cost volume is implemented in CUDA with PyTorch interface.

\begin{table}[t!]
	\begin{center}
    \begin{scriptsize}
    
			\begin{tabular}{|l|c|c|c|c|}
				\hline
				 &  \multicolumn{2}{c|}{Train}  & \multicolumn{2}{c|}{Test}    \\  \cline{2-5} 
				 & Clean & Final & Clean & Final  \\    
				\hline            
				EpicFlow \cite{epicflow} & - & - & 4.12 & 6.29 \\
				MRFlow \cite{mrflow}   & 1.83 & 3.59 & 2.53 & 5.38 \\
				FlowFields \cite{bailer2015iccvflowfields} & - & - & 3.78 & 5.36  \\
				DCFlow \cite{XuCVPR2017DCFlow}   & - & - & 3.54 & 5.12 \\            
				\hline
				FlowNetS  & 4.35 & 5.46 &-&-  \\ 
				FlowNetS (ft) & (3.66) & (4.44) & 6.96 & 7.52  \\
				FlowNetC & 3.52 & 5.00 & - & - \\
				FlowNetC (ft) &  (3.50) & (3.89) & 6.85 & 8.51 \\
				FlowNet2 & \textbf{2.02} & \textbf{3.14} & \textbf{3.96} & 6.02  \\
				FlowNet2 (ft)  & (1.45)& (2.01) &4.16 &5.74 \\
				SpyNet & 4.12 & 5.57  & 6.69 & 8.43  \\
				SpyNet (ft)  & (3.17)& (4.32)& 6.64 & 8.36 \\
				PWC-Net & 2.55 &  3.93 & -  & -  \\
				PWC-Net (ft) & (2.02) & (2.08) & 4.39 & \textbf{5.04} \\
				LiteFlowNet  & 2.48 &  4.04 & - & -  \\
				LiteFlowNet (ft) & (1.35) & (1.78) & 4.54 & 5.38 \\
				Devon & 2.45 & 3.72 & - & -   \\
				Devon (ft) & (1.97)& (2.67)  & 4.34 &  6.35 \\
				\hline
			\end{tabular}               
    \end{scriptsize}
	\end{center}
	\caption{Results on Sintel (end-point error). (ft) denotes the fine-tuning. }
	\label{table:sintel}
\end{table}

\begin{table}[h!]
	\begin{center}
    \begin{scriptsize}
    
\begin{tabular}{|l|c|c|c|c|c|c|}
\hline
Clean & d0-10 & d10-60 & d60-140 & s0-10 & s10-40 & s40+ \\
\hline
SpyNet & 6.69 &4.37 &3.29 &1.40 &5.53 &49.71 \\
FlowNet2  & 4.82 & 2.56 & 1.74 & 0.96 & 3.23 & 35.54 \\
PWC-Net & 4.68 & 2.08 & 1.52 & 0.90 & 2.99 & 31.28 \\
LiteFlowNet & \textbf{3.27} & \textbf{1.44} &	0.93 & 0.50 & \textbf{1.73} & 31.41 \\
Devon & 4.12 & 1.53 & \textbf{0.82} & \textbf{0.76} & 2.45 & \textbf{26.72}  \\
\hline
Final & d0-10 & d10-60 & d60-140 & s0-10 & s10-40 & s40+ \\
\hline
SpyNet & 5.50 &3.12 &1.72 &0.83 &3.34 &43.44 \\
FlowNet2  & \textbf{3.27} &1.46 &0.86 &\textbf{0.60} & \textbf{1.89} &27.35 \\
PWC-Net & 3.83 & \textbf{1.31} & \textbf{0.56} & 0.70& 2.19& \textbf{23.56} \\
LiteFlowNet & 4.09 & 2.10 & 1.73 & 0.75	& 2.75 &	34.72 \\
Devon & 5.34 & 2.88 & 2.30 & 1.12 & 3.83 & 38.38 \\
\hline
\end{tabular}               
\end{scriptsize}
\end{center}
\caption{Detailed results on Sintel (end-point error) for different distances from motion boundaries (d) and 
velocities (s).}
\label{table:sintel_large}
\end{table}

\begin{table}[h!]
	\begin{center}
    \begin{scriptsize}
    
\begin{tabular}{|l|c|c|c|c|}
\hline
 & \multicolumn{2}{c|}{KITTI 2012} & \multicolumn{2}{c|}{KITTI 2015}  \\  \cline{2-5} 
 & Train & Test & Train & Test \\
 &  EPE & EPE & EPE & F1-all\\
\hline            
EpicFlow &  3.09 &   3.8 & 27.18\% & 27.10\% \\
MRFlow    & - & - & 14.09\% & 12.19\% \\
FlowFields & - & 3.0 & - & 19.80\%  \\
DCFlow   & - & - & 15.09\% & 14.83\% \\  
\hline
FlowNetS  & 8.26 &- &- &-  \\ 
FlowNetS (ft)  & - & 9.1 &- &- \\
FlowNetC & 9.35 &-&-&- \\
FlowNetC (ft) & -&-&-&- \\		
FlowNet2 & 4.09 & -  & \textbf{10.08} &- \\				
FlowNet2 (ft)  & (1.28) & 1.8 &  (2.30) & 10.41\% \\
SpyNet & 9.12 &-&-&-  \\
SpyNet (ft) & (4.13) & 4.7 & - & 35.07\%  \\
PWC-Net & 4.14 & - &  10.35 & - \\
PWC-Net (ft) & (1.08) & 1.7  & (2.16) &\textbf{ 9.16\%} \\
LiteFlowNet  & \textbf{4.00} & - & 10.39 & -  \\
LiteFlowNet (ft) & (1.05) & \textbf{1.6}  & (1.62) & 9.38\%  \\
Devon &  4.73 & - & 10.65 & -  \\
Devon (ft) & (1.29) & 2.6 & (2.00) & 14.31 \% \\
\hline
			\end{tabular}               
    \end{scriptsize}
	\end{center}
	\caption{Results on KITTI. (ft) denotes the fine-tuning. EPE denotes end-point error.  Fl-all denotes the ratio of pixels where the flow estimate is incorrect by both $\geq$ 3 pixels and $\geq$ 5\%. }
	\label{table:kitti}
\end{table}

\subsection{Main Results}
\begin{figure*}[h]
\centering
\subfloat[First image]{
\includegraphics[width=0.25\textwidth]{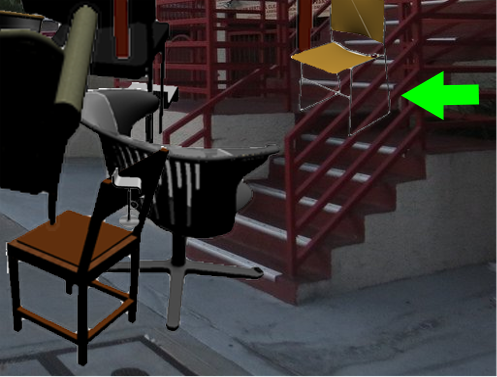}}
\subfloat[Second image]{
\includegraphics[width=0.25\textwidth]{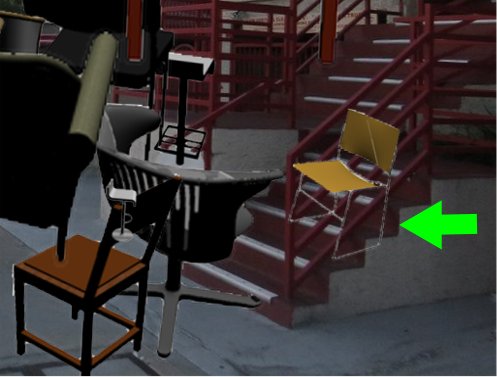}}
\subfloat[Ground truth]{
\frame{\includegraphics[width=0.25\textwidth]{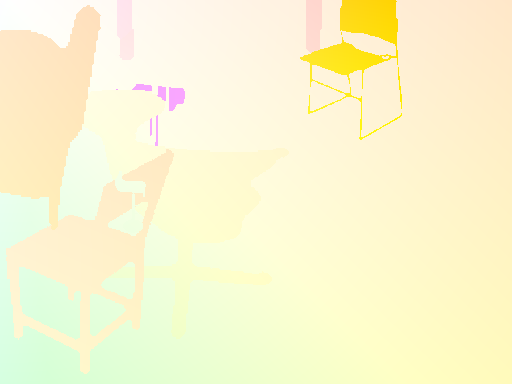}}}

\subfloat[LiteFlowNet]{
\frame{\includegraphics[width=0.25\textwidth]{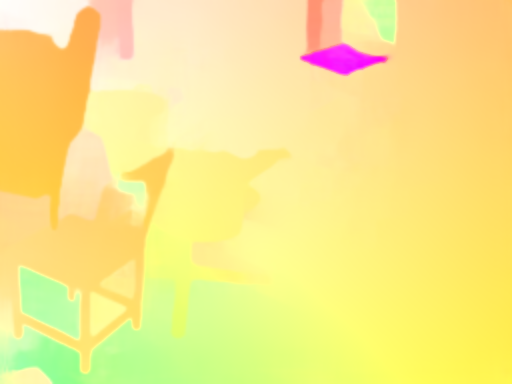}}}
\subfloat[PWC-Net]{
\frame{\includegraphics[width=0.25\textwidth]{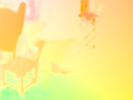}}}
\subfloat[Devon]{
\frame{\includegraphics[width=0.25\textwidth]{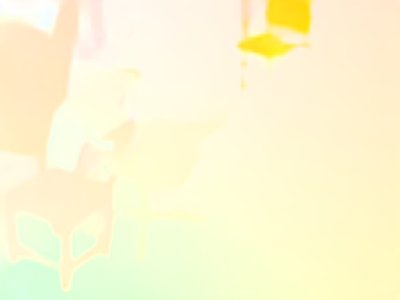}}}
\vspace{0.2cm}
\caption{FlyingChairs (validation set). Green arrows indicate the small object that moves fast.}        
\label{fig:fc}
\end{figure*}

\begin{figure*}[h]
\centering
\subfloat[First image]{
\includegraphics[width=0.3\textwidth]{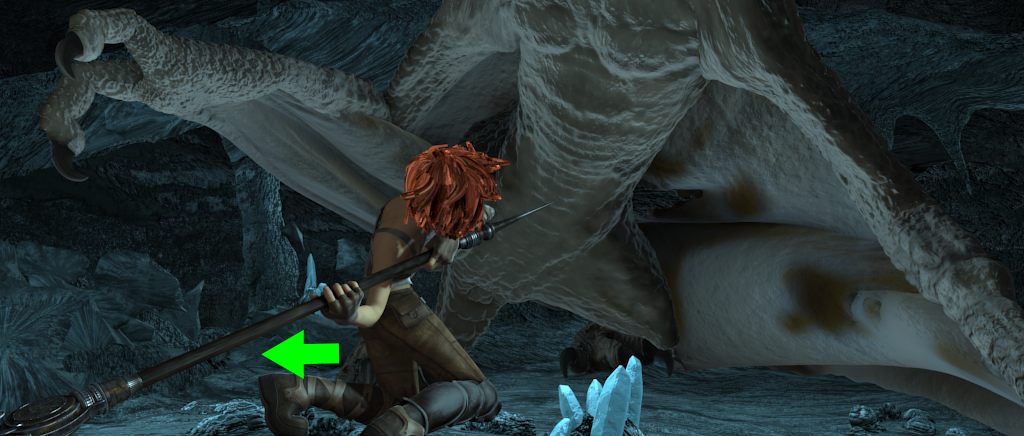}}
\subfloat[Second image]{
\includegraphics[width=0.3\textwidth]{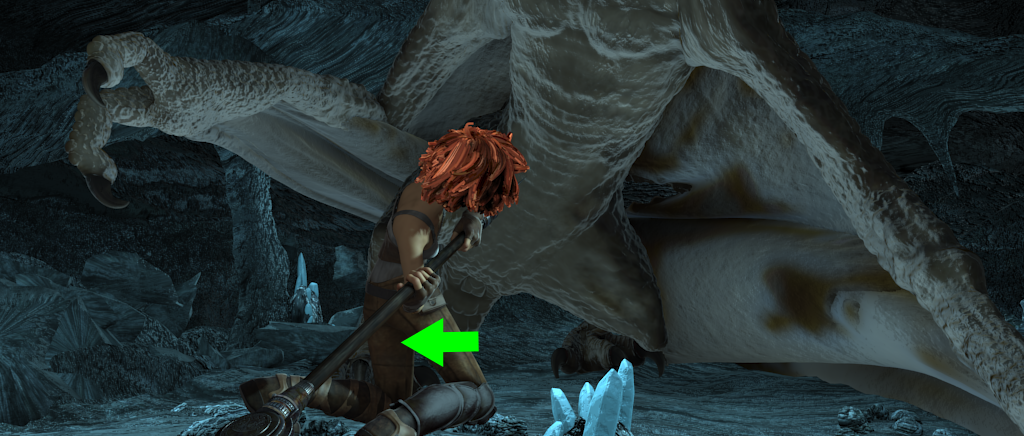}}
\subfloat[Ground truth]{
\frame{\includegraphics[width=0.3\textwidth]{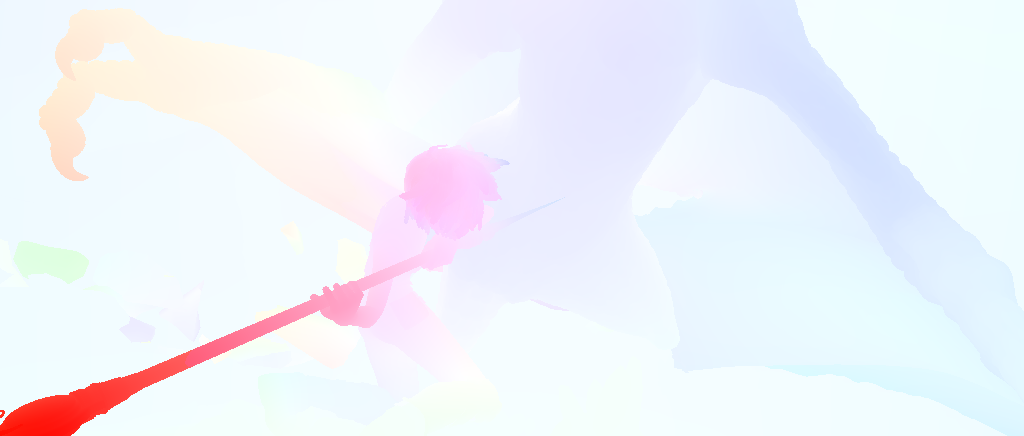}}}

\subfloat[LiteFlowNet]{
\frame{\includegraphics[width=0.3\textwidth]{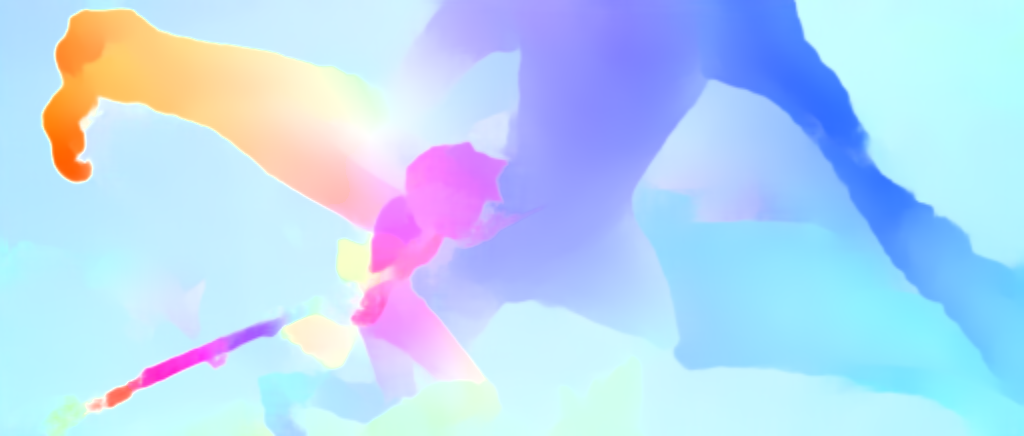}}}
\subfloat[PWC-Net]{
\frame{\includegraphics[width=0.3\textwidth]{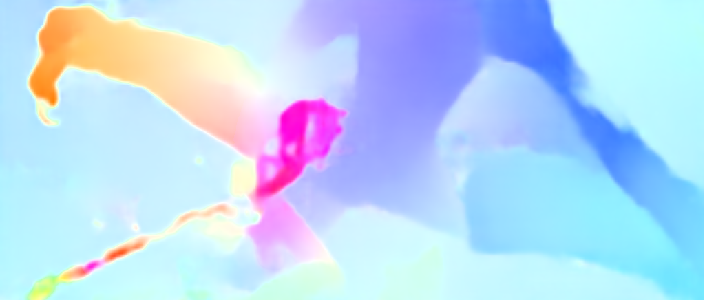}}}
\subfloat[Devon]{
\frame{\includegraphics[width=0.3\textwidth]{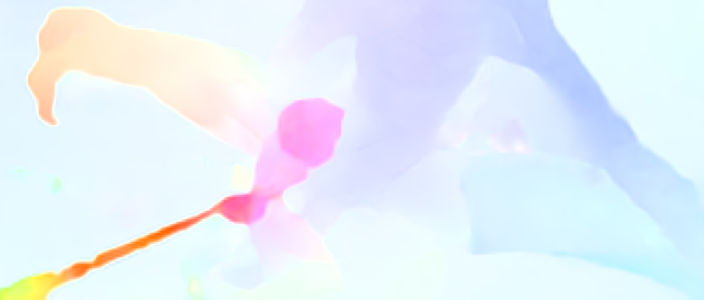}}}
\vspace{0.2cm}
\caption{Sintel (training set). Green arrows indicate the small object that moves fast.}        
\label{fig:sintel}
\end{figure*}

\begin{figure*}[h]
\centering
\subfloat[First image]{
\includegraphics[width=0.3\textwidth]{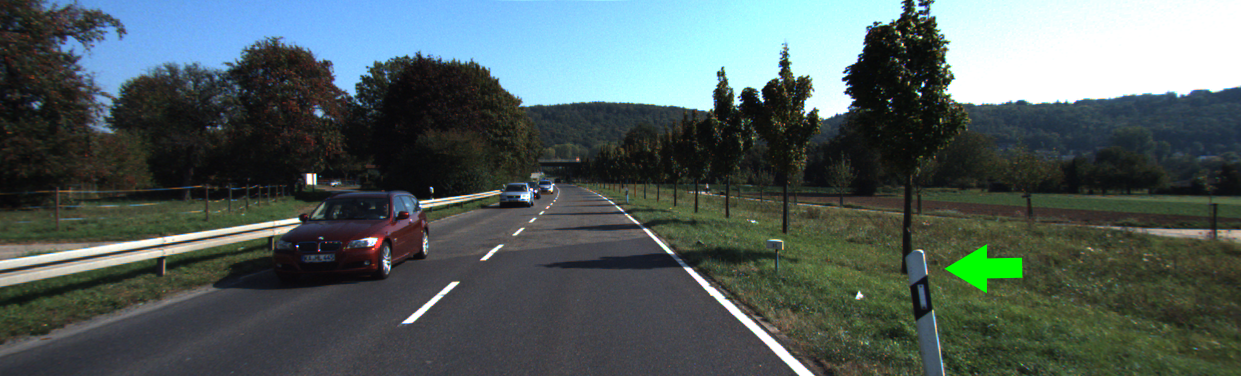}}
\subfloat[Second image]{
\includegraphics[width=0.3\textwidth]{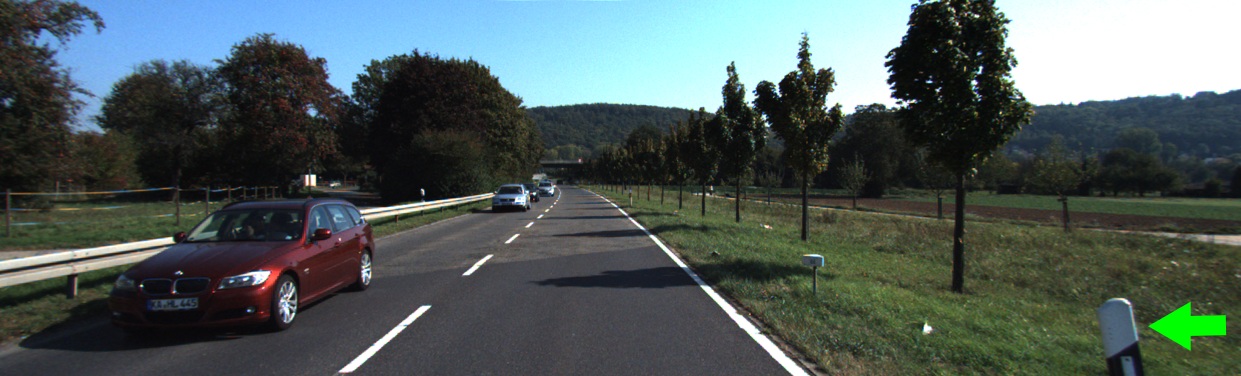}}
\subfloat[Ground truth]{
\frame{\includegraphics[width=0.3\textwidth]{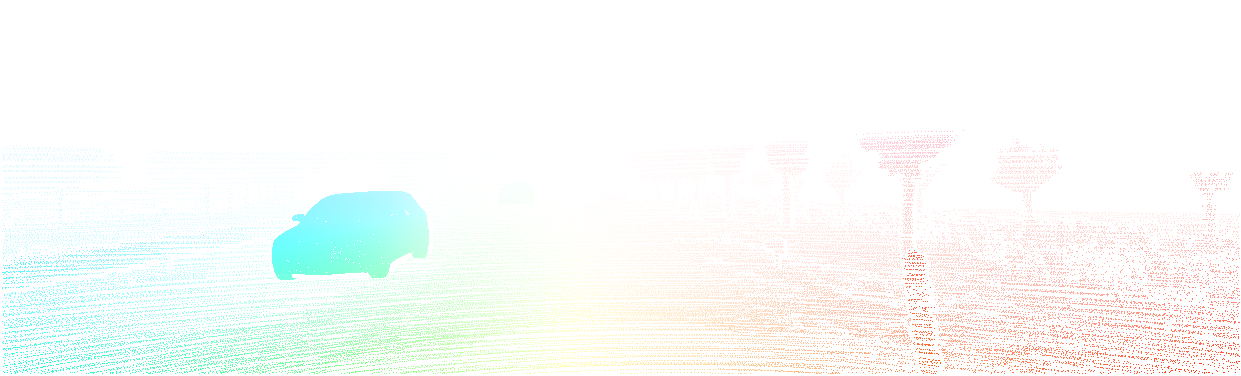}}}

\subfloat[LiteFlowNet]{
\frame{\includegraphics[width=0.3\textwidth]{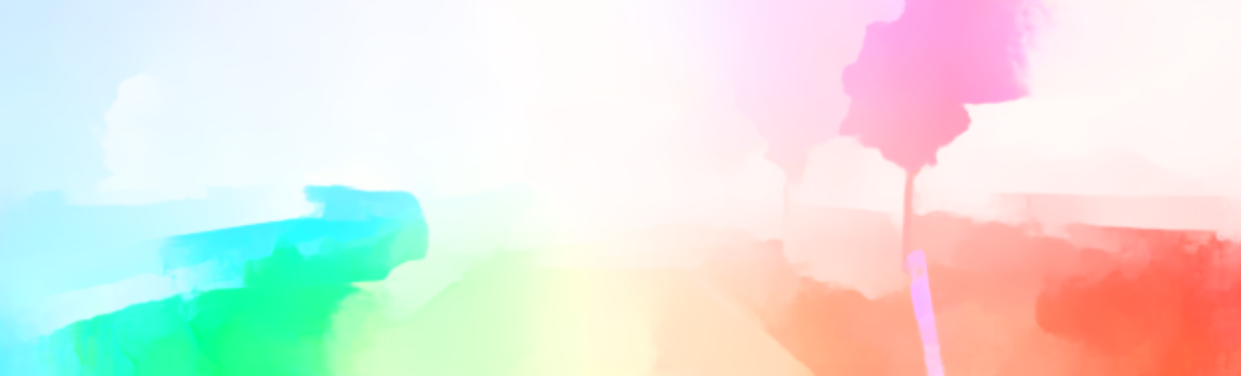}}}
\subfloat[PWC-Net]{
\frame{\includegraphics[width=0.3\textwidth]{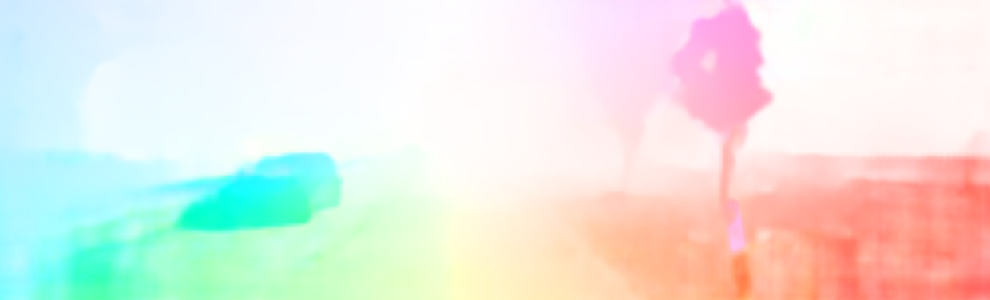}}}
\subfloat[Devon]{
\frame{\includegraphics[width=0.3\textwidth]{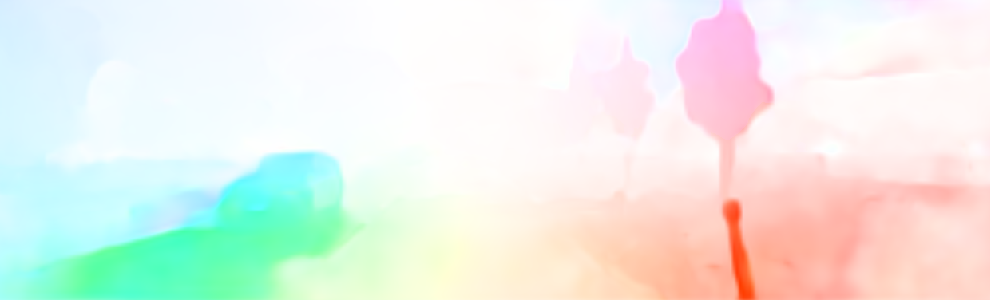}}}
\vspace{0.2cm}
\caption{KITTI 2015 (training set). Green arrows indicate the small object that moves fast.}        
\label{fig:kitti}
\end{figure*}

In Figure \ref{fig:fc}, \ref{fig:sintel} and \ref{fig:kitti}, we show visualization results of situations where small objects move fast. All models were trained on FlyingChairs and then fine-tuned on FlyingThings3D. No additional fine-tuning is applied.

\clearpage

\clearpage

From Figure \ref{fig:fc}, \ref{fig:sintel} and \ref{fig:kitti}, we can see Devon gives much more accurate estimation of the small objects. 
The results on Sintel and KITTI are listed in Table \ref{table:sintel}, \ref{table:sintel_large} and \ref{table:kitti}, from which we can see Devon outperforms PWC-Net and LiteFlowNet on Sintel clean pass, though not on Sintel final pass and KITTI.

\subsection{Ablation Analysis}\label{sec:ablation}

We perform an ablation analysis of Devon trained on FlyingChairs. There are six ablation cases: (1) With warping. We replace the deformable cost volumes with warping and standard cost volumes (with dilation). (2) With shortcut. We additionally concatenate the relation module with the feature maps of the first image and feed them into the decoder. (3) Without dilation. We replace the concatenated deformable cost volumes with one deformable cost volume of neighorhood size 13$\times$ 13 and dilation rate one. (4) Without norm. We remove the normalization in the relation modules. (5) Shared decoder. We let all stages share one decoder and set the hyparameters of the relation modules in second and third stage the same as the first stage. (6) Simple encoder. We replace the U-Net structure encoding module with a  simpler one: 4 convolutional layers of 32 units receptive field size 3$\times$3. The first two layers have stride 2 and the last two have stride 1. For (1)$\sim$(5), the changes are applied to all stages in Devon.
The results are listed in Table \ref{table:ablation}, from which we can see the architecture of Devon is robust to various changes.

\begin{table}[h!]
	\begin{center}
    \begin{small}
		\begin{tabular}{|l|c|c|c|}
			\hline
				 &  FlyingChairs & Sintel clean   &  KITTI 2015 \\  
			& Valid  & Train & Train   \\
			\hline
			Full model & 1.87 &2.99  &  \textbf{13.25} \\
			With warping & 1.88 & 2.98  & 13.73  \\
			With shortcut & \textbf{1.79} & \textbf{2.97} & 15.31 \\
			Without dilation  & 1.95 & 3.12 &  13.84 \\
			Without norm  & 1.99 & 3.30  &  15.64 \\
			Shared decoder & 1.96 & 2.90  &  13.89 \\
			Simple encoder & 1.88 &  3.00  & 14.04\\
			\hline
		\end{tabular}    
    \end{small}
	\end{center}
	\caption{Results of ablation experiments after training on FlyingChairs (end-point error).}
    \label{table:ablation}
\end{table}

\begin{table}[h!]
	\begin{center}
    \begin{small}
		\begin{tabular}{|l|c|c|}
			\hline
				 &  Forward & Backward  \\
			\hline
			Full model   & 50.51 & 177.17\\
			With warping   & 57.75 & 182.75 \\ 
			With shortcut & 51.62 & 181.47 \\
			Without dilation & \textbf{49.07} & \textbf{147.74} \\
			Without norm & 52.33 & 177.25 \\
			Shared decoder & 51.15 & 178.54 \\
			Simple encoder & 49.98 & 180.78 \\
			\hline
		\end{tabular}    
    \end{small}
	\end{center}
	\caption{Runtime (ms).}
    \label{table:runtime}
\end{table}

\subsection{Runtime}
We report the runtime of Devon and its variants in Table \ref{table:runtime}. The timing was recorded on a NVIDIA TITAN Xp graphics card for processing a pair of RGB images of size $1024\times 448$. 

\section{Discussions}

While Devon achieves better results in handling small objects moving fast in the visualization results, it does not outperform multi-resolution based methods such as PWC-Net on Sintel final pass and KITTI. We conjecture that this is due to the fact that Sintel and KITTI mostly contain large objects (e.g. human bodies, cars and buildings), for which the multi-resolution approach might be more suitable. An interesting extension of our work is to combine multi-resolution approach and Devon to handle objects of diverse sizes and speed.

From Table \ref{table:runtime}, we can see using deformable cost volume achieves shorter runtime than using warping and standard cost volumes. This is because the deformation and the cost volume construction are merged into one process in deformable cost volume and therefore reduces the runtime. Along with Table \ref{table:runtime}, the result suggests that the deformable cost volume is a suitable replacement of warping and standard cost volume in Devon and potentially other models.

The deformation in deformable cost volume is different from the one in deformable convolutional networks \cite{dai17dcn}. In deformable cost volume, the cost volume is offset by an external optical flow and dilation. There is no learnable parameter while in deformable convolutional networks, the deformation is element-wise and the offset parameters are learned during training. Another related work is the deformable spatial pyramid matching \cite{kim2013deformable,hur2015generalized} which uses deformation in the classic energy minimization framework for hierarchical dense matching. Applying  normalization on hidden unit outputs is found advantageous in modeling general image relations  \cite{lu2017learning}.

The use of dilation in standard cost volume has been proposed in \cite{fischer2015flownet} and used in \cite{hui2018liteflownet}, though it has not been used in a multi-scale fashion in each stage as ours to handle the small objects moving fast problem.

\section{Conclusions}
In this paper, we proposed a new neural network module, Deformable Cost Volume, which allows the representation of multi-scale motion in a single high resolution and avoids the drawback of warping. Based on it, we designed the Deformable Volume Network, which is demonstrated to be effective in estimating optical flow, especially in situations where small objects move fast.

\clearpage

{\small
	\bibliographystyle{ieee}
	\bibliography{devon}
}

\end{document}

%% file: img/scv.tex
	\begin{tikzpicture}
	\draw[step=0.2,gray!20,thin] (0,0) grid (2.4,2.4);
	\draw[step=0.2,gray!20,thin] (3,0) grid (5.4,2.4);
    \node[align=center] at (1.2,-0.3) {${f}_I$};
    \node[align=center] at (4.2,-0.3) {${f}_J$};
					
	\draw[-,thick] (0,0) -- (0,2.4);
	\draw[-,thick] (0,0) -- (2.4,0);
	\draw[-,thick] (2.4,0) -- (2.4,2.4);
	\draw[-,thick] (0,2.4) -- (2.4,2.4);
	
	\draw[-,thick] (3,0) -- (3,2.4);
	\draw[-,thick] (3,0) -- (5.4,0);
	\draw[-,thick] (5.4,0) -- (5.4,2.4);
	\draw[-,thick] (3,2.4) -- (5.4,2.4);
	
	\filldraw[fill=blue!40!white, draw=black] (0.6,1.6) rectangle (0.8,1.4);		
	\filldraw[fill=blue!40!white, draw=black] 
	(3.4,1.2) rectangle (3.6,1.4);
	\filldraw[fill=blue!40!white, draw=black] 
	(3.6,1.2) rectangle (3.8,1.4);
	\filldraw[fill=blue!40!white, draw=black] 
	(3.8,1.2) rectangle (4.0,1.4);
	\filldraw[fill=blue!40!white, draw=black] 
	(3.4,1.4) rectangle (3.6,1.6);
	\filldraw[fill=blue!40!white, draw=black] 
	(3.6,1.4) rectangle (3.8,1.6);
	\filldraw[fill=blue!40!white, draw=black] 
	(3.8,1.4) rectangle (4.0,1.6);
	\filldraw[fill=blue!40!white, draw=black] 
	(3.4,1.6) rectangle (3.6,1.8);
	\filldraw[fill=blue!40!white, draw=black] 
	(3.6,1.6) rectangle (3.8,1.8);
	\filldraw[fill=blue!40!white, draw=black] 
	(3.8,1.6) rectangle (4.0,1.8);				
	
	\filldraw[fill=green!40!white] (1.4,0.8) rectangle (1.6,1.0);		
	\filldraw[fill=green!40!white] 
	(4.2,1.0) rectangle (4.4,1.2);
	\filldraw[fill=green!40!white] 
	(4.4,1.0) rectangle (4.6,1.2);
	\filldraw[fill=green!40!white] 
	(4.6,1.0) rectangle (4.8,1.2);
	\filldraw[fill=green!40!white] 
	(4.2,0.8) rectangle (4.4,1.0);
	\filldraw[fill=green!40!white] 
	(4.4,0.8) rectangle (4.6,1.0);
	\filldraw[fill=green!40!white] 
	(4.6,0.8) rectangle (4.8,1.0);
	\filldraw[fill=green!40!white] 
	(4.2,0.6) rectangle (4.4,0.8);
	\filldraw[fill=green!40!white] 
	(4.4,0.6) rectangle (4.6,0.8);
	\filldraw[fill=green!40!white] 
	(4.6,0.6) rectangle (4.8,0.8);
	\end{tikzpicture}

%% file: img/dcv.tex
	\begin{tikzpicture}[]
	\draw[step=0.2,gray!20,thin] (0,0) grid (2.4,2.4);
	\draw[step=0.2,gray!20,thin] (3,0) grid (5.4,2.4);
    \node[align=center] at (1.2,-0.3) {${f}_I$};
    \node[align=center] at (4.2,-0.3) {${f}_J$};		
	
	\draw[-,thick] (0,0) -- (0,2.4);
	\draw[-,thick] (0,0) -- (2.4,0);
	\draw[-,thick] (2.4,0) -- (2.4,2.4);
	\draw[-,thick] (0,2.4) -- (2.4,2.4);
	
	\draw[-,thick] (3,0) -- (3,2.4);
	\draw[-,thick] (3,0) -- (5.4,0);
	\draw[-,thick] (5.4,0) -- (5.4,2.4);
	\draw[-,thick] (3,2.4) -- (5.4,2.4);

	\filldraw[fill=blue!40!white, draw=black] (0.6,1.6) rectangle (0.8,1.4);	
		
	\filldraw[fill=blue!40!white, draw=black] 
	(3.8,1.2) rectangle (4.0,1.4);
	\filldraw[fill=blue!40!white, draw=black] 
	(4.2,1.2) rectangle (4.4,1.4);
	\filldraw[fill=blue!40!white, draw=black] 
	(4.6,1.2) rectangle (4.8,1.4);
	\filldraw[fill=blue!40!white, draw=black] 
	(3.8,1.6) rectangle (4.0,1.8);
	\filldraw[fill=blue!40!white, draw=black] 
	(4.2,1.6) rectangle (4.4,1.8);
	\filldraw[fill=blue!40!white, draw=black] 
	(4.6,1.6) rectangle (4.8,1.8);
	\filldraw[fill=blue!40!white, draw=black] 
	(3.8,2.0) rectangle (4.0,2.2);
	\filldraw[fill=blue!40!white, draw=black] 
	(4.2,2.0) rectangle (4.4,2.2);
	\filldraw[fill=blue!40!white, draw=black] 
	(4.6,2.0) rectangle (4.8,2.2);				
	
	\filldraw[fill=green!40!white] (1.4,0.8) rectangle (1.6,1.0);		
	\filldraw[fill=green!40!white] 
	(3.4,1.0) rectangle (3.6,1.2);
	\filldraw[fill=green!40!white] 
	(3.8,1.0) rectangle (4.0,1.2);
	\filldraw[fill=green!40!white] 
	(4.2,1.0) rectangle (4.4,1.2);
	\filldraw[fill=green!40!white] 
	(3.4,0.6) rectangle (3.6,0.8);
	\filldraw[fill=green!40!white] 
	(3.8,0.6) rectangle (4.0,0.8);
	\filldraw[fill=green!40!white] 
	(4.2,0.6) rectangle (4.4,0.8);
	\filldraw[fill=green!40!white] 
	(3.4,0.2) rectangle (3.6,0.4);
	\filldraw[fill=green!40!white] 
	(3.8,0.2) rectangle (4.0,0.4);
	\filldraw[fill=green!40!white] 
	(4.2,0.2) rectangle (4.4,0.4);

	\draw[-latex,thick] (0.7,1.5) -- (1.3,1.7);		
	\draw[-latex,thick] (3.7,1.5) -- (4.3,1.7);				
	
	\draw[-latex,thick] (1.5,0.9) -- (0.9,0.7);		
	\draw[-latex,thick] (4.5,0.9) -- (3.9,0.7);
	
	\end{tikzpicture}

%% file: img/devon_archi.tex
	\begin{tikzpicture}[auto,node distance=0.5cm,thick]
	
	\node [auto,align=center,thick,draw,fill=green!20,minimum width=1cm,minimum height=0.75cm] at (0,0)(I)  {\scriptsize{$J$}}; 
	
	\node [auto,align=center,thick,draw,fill=green!20,minimum width=1cm,minimum height=0.75cm,,left=of I] (J) {\scriptsize{$I$}}; 	
	
	\node [module,below=of I] (f_I) 
	{\scriptsize{$f$}}; 		
	\node [module,below=of J] (f_J) 
	{\scriptsize{$f$}}; 
	
	\node [module,draw,right=of f_I,yshift=-0.75cm] (R1)  {\scriptsize{$R_1$}};
	\node [module,draw,right=of R1,xshift=0.5cm] (g1){\scriptsize{$g_1$}};
	
	\node [module,draw,below=of R1,yshift=-0.5cm] (R2)  {\scriptsize{$R_2$}};
	\node [module,draw,right=of R2,xshift=0.5cm] (g2){\scriptsize{$g_2$}};
	
	\node [module,draw,below=of R2,yshift=-0.5cm] (R3)  {\scriptsize{$R_3$}};
	\node [module,draw,right=of R3,xshift=0.5cm] (g3){\scriptsize{$g_3$}};
	
	\node [right=of g1] (a1) {};	
	\node [fill=white!20,draw,circle,inner sep=0.01cm,minimum width=0.05cm,minimum height=0.05cm,right=of g2] (a2)  {\scriptsize{$+$}}; 
	\node [fill=white!20,draw,circle,inner sep=0.01cm,minimum width=0.05cm,minimum height=0.05cm,right=of g3] (a3)  {\scriptsize{$+$}}; 

	\node [fill=blue!10,draw,minimum width=0.67cm,minimum height=0.5cm,right=of a2] (F2)  {\scriptsize{$F_2$}}; 	
	\node [fill=blue!10,draw,minimum width=0.67cm,minimum height=0.5cm,right=of a3] (F3)  {\scriptsize{$F_3$}};
	\node [fill=blue!10,draw,minimum width=0.67cm,minimum height=0.5cm] (F1) at (g1-|F2)  {\scriptsize{$F_1$}};  	
	
    \coordinate[right=of R1,yshift=0.1cm] (j1_B1) {};
	\coordinate[right=of R1,yshift=-0.1cm] (j1_B2) {};	
    \coordinate[right=of R2,yshift=0.1cm] (j2_B1) {};
	\coordinate[right=of R2,yshift=-0.1cm] (j2_B2) {};	
    \coordinate[right=of R3,yshift=0.1cm] (j3_B1) {};
	\coordinate[right=of R3,yshift=-0.1cm] (j3_B2) {};		
	
	\coordinate[yshift=0.1cm] (j1_I) at (I|-R1) {};
	\coordinate[yshift=-0.1cm] (j1_J) at (J|-R1){};
	\coordinate[yshift=0.1cm] (j2_I) at (I|-R2) {};
	\coordinate[yshift=-0.1cm] (j2_J) at (J|-R2){};
	\coordinate[yshift=0.1cm] (j3_I) at (I|-R3) {};
	\coordinate[yshift=-0.1cm] (j3_J) at (J|-R3){};	
	\coordinate (j1_a1) at ($(F1)!0.5!(F2)$) {};
	\coordinate (j1_a2) at (j1_a1-|a2) {};
	\coordinate (j1_a3) at (j1_a1-|R1) {};
	
	\coordinate (j2_a1) at ($(F2)!0.5!(F3)$) {};
	\coordinate (j2_a2) at (j2_a1-|a2) {};
	\coordinate (j2_a3) at (j2_a1-|R2) {};

	\draw[-latex] (I) -- (f_I);	
	\draw[-latex] (J) -- (f_J);
	\draw[-] (f_I) -- (j1_I);
	\draw[-] (f_J) -- (j1_J);
	\draw[-] (j1_I) -- (j2_I);
	\draw[-] (j1_J) -- (j2_J);	
	\draw[-] (j2_I) -- (j3_I);
	\draw[-] (j2_J) -- (j3_J);		
	\draw[-latex] (j1_I) -- (j1_I-|R1.west);
	\draw[-latex] (j1_J) -- (j1_J-|R1.west);
	\draw[-latex] (j2_I) -- (j2_I-|R2.west);
	\draw[-latex] (j2_J) -- (j2_J-|R2.west);
	\draw[-latex] (j3_I) -- (j3_I-|R3.west);
	\draw[-latex] (j3_J) -- (j3_J-|R3.west);

	\draw[-latex] (R1) -- (g1);
	\draw[-latex] (R2) -- (g2);	
    \draw[-latex] (R3) -- (g3);
	\draw[-latex] (g1) -- (F1);
	\draw[-latex] (g2) -- (a2);
	\draw[-latex] (a2) -- (F2);
	\draw[-latex] (g3) -- (a3);
	\draw[-latex] (a3) -- (F3);
	
	\draw[-] (F1) -- (j1_a1);
	\draw[-] (j1_a1) -- (j1_a2);
	\draw[-] (j1_a2) -- (j1_a3);
	\draw[-latex] (j1_a2) -- (a2);
	\draw[-latex] (j1_a3) -- (R2);
	
	\draw[-] (F2) -- (j2_a1);
	\draw[-] (j2_a1) -- (j2_a2);
	\draw[-] (j2_a2) -- (j2_a3);
	\draw[-latex] (j2_a2) -- (a3);
	\draw[-latex] (j2_a3) -- (R3);

	\node[fill,circle,inner sep=0pt,minimum size=0.1cm] at (j1_a1-|a2) {};
	\node[fill,circle,inner sep=0pt,minimum size=0.1cm] at (j2_a1-|a2) {};	

	\node[fill,circle,inner sep=0pt,minimum size=0.1cm,yshift=0.1cm] at (I|-R1) {};
	\node[fill,circle,inner sep=0pt,minimum size=0.1cm,yshift=0.1cm] at (I|-R2) {};
	\node[fill,circle,inner sep=0pt,minimum size=0.1cm,yshift=-0.1cm] at (J|-R1) {};	
	\node[fill,circle,inner sep=0pt,minimum size=0.1cm,yshift=-0.1cm] at (J|-R2) {};	
	
	\end{tikzpicture}

%% file: img/f_rich.tex
	\begin{tikzpicture}[auto,node distance=0.35cm,thick]			

	\node [module,minimum width=3.2cm] (c1) at (0,0) 
	{\scriptsize{Conv $16\times 3\times 3$ , stride 2}}; 
	\node [module,minimum width=3.2cm,below=of c1] (c2)  
	{\scriptsize{Conv $32\times 3\times 3$, stride 2}}; 
	\node [module,minimum width=3.2cm,below=of c2] (c3)  
	{\scriptsize{Conv $64\times 3\times 3$, stride 2}}; 
	\node [module,minimum width=3.2cm,below=of c3] (c4)  
	{\scriptsize{Conv $128\times 3\times 3$, stride 2}}; 
	\node [module,minimum width=3.2cm,below=of c4] (c5)  
	{\scriptsize{Conv $256\times 3\times 3$, stride 2}}; 	
	\node [module,minimum width=3.2cm,below=of c5] (c6)
	{\scriptsize{Conv $512\times 3\times 3$, stride 2}}; 	
	\node [module,minimum width=3.2cm,below=of c6] (c7)  	
	{\scriptsize{Conv $512\times 3\times 3$, stride 1}}; 	
	\node [module,minimum width=3.2cm,below=of c7] (c8)
	{\scriptsize{Conv $256\times 3\times 3$, stride 1}};
	\node [module,minimum width=3.2cm,below=of c8] (c9)
	{\scriptsize{Conv $128\times 3\times 3$, stride 1}}; 		
	\node [module,minimum width=3.2cm,below=of c9] (c10)  	
	{\scriptsize{Conv $64\times 3\times 3$, stride 1}}; 		
	\node [module,minimum width=3.2cm,below=of c10] (c11)  
	{\scriptsize{Conv $32\times 3\times 3$, stride 1}};
	\node [above=of c1] (c0)
	{\scriptsize{Image}}; 		
	
	\node[left=of c1,xshift=-1.5cm] (l)  {};	
	\node[below=of c11] (output) {\scriptsize{Image Features}};	        	
	\draw[-latex] (c0) -- (c1);	
	\draw[-latex] (c1) -- (c2);	
	\draw[-latex] (c2) -- (c3);					    	
	\draw[-latex] (c3) -- (c4);					    			
	\draw[-latex] (c4) -- (c5);		
    \draw[-latex] (c5) -- (c6);
    \draw[-latex] (c6) -- (c7);
	\draw[-latex] (c7) edge node[right] {\scriptsize{Upsample $\times 2$}} (c8);	
	\draw[-latex] (c8) edge node[right] {\scriptsize{Upsample $\times 2$}} (c9);	
	\draw[-latex] (c9) edge node[right] {\scriptsize{Upsample $\times 2$}} (c10);
	\draw[-latex] (c10) edge node[right] {\scriptsize{Upsample $\times 2$}} (c11);
	\draw[-latex] (c11) -- (output);

	\draw [-latex] (c6.east) to [bend right=-90] (c7.east);
	\draw [-latex] (c5.east) to [bend right=-90] (c8.east);
	\draw [-latex] (c4.east) to [bend right=-90] (c9.east);
	\draw [-latex] (c3.east) to [bend right=-90] (c10.east);
	\draw [-latex,postaction={decorate,decoration={raise=0.1cm,text along path,text align=center,text={|\scriptsize|Residual Connection}}}] (c2.east) to [bend right=-90]  (c11.east);	
	
	\end{tikzpicture}

%% file: img/R.tex
\begin{tikzpicture}[auto,node distance=0.4cm,thick]		
\node [module] (dcv1)  at (-3.4,0) 
{\scriptsize{$C_1$, $k_1$, $r_1$}};
\node [module] (dcv2) at (-1.7,0) 
{\scriptsize{$C_2$, $k_2$, $r_2$}}; 	
\node [module] (dcv3)  at (0,0) 
{\scriptsize{$C_3$, $k_3$, $r_3$}}; 	
\node [module] (dcv4)  at (1.7,0) 
{\scriptsize{$C_4$, $k_4$, $r_4$}};
\node [module] (dcv5)  at (3.4,0) 
{\scriptsize{$C_5$, $k_5$, $r_5$}}; 
\node [module,below=of dcv3] (concat)  
{\scriptsize{Concat}}; 	
\node [module,below=of concat] (norm)  
{\scriptsize{Norm}}; 		

\node[] (flow) at (-2,1)  {\scriptsize{Image Features 1}};    	
\node[] (input1) at (0,1)  {\scriptsize{Flow}};	    
\node[] (input2) at (2,1)  {\scriptsize{Image Features 2}};	       	

\node[below=of norm] (output) {\scriptsize{Relation Features}};	  

\draw[-latex] (input1.south) -- (dcv1.north);
\draw[-latex] (input1.south) -- (dcv2.north);
\draw[-latex] (input1.south) -- (dcv3.north);
\draw[-latex] (input1.south) -- (dcv4.north);
\draw[-latex] (input1.south) -- (dcv5.north);
\draw[-latex] (input2.south) -- (dcv1.north);
\draw[-latex] (input2.south) -- (dcv2.north);
\draw[-latex] (input2.south) -- (dcv3.north);
\draw[-latex] (input2.south) -- (dcv4.north);
\draw[-latex] (input2.south) -- (dcv5.north);
\draw[-latex] (flow.south) -- (dcv1.north);
\draw[-latex] (flow.south) -- (dcv2.north);
\draw[-latex] (flow.south) -- (dcv3.north);
\draw[-latex] (flow.south) -- (dcv4.north);
\draw[-latex] (flow.south) -- (dcv5.north);
\draw[-latex] (dcv1.south) -- (concat.north);					        
\draw[-latex] (dcv2.south) -- (concat.north);
\draw[-latex] (dcv3.south) -- (concat.north);
\draw[-latex] (dcv4.south) -- (concat.north);		
\draw[-latex] (dcv5.south) -- (concat.north);		
\draw[-latex] (concat.south) --(norm.north);	
\draw[-latex] (norm.south) --(output.north);				
\end{tikzpicture}

%% file: img/retina.tex
	\begin{tikzpicture}
	\begin{scope}[scale=0.8, transform shape]
	\draw[step=0.2,gray!20,thin] (0,0) grid (3.0,3.0);
	\draw[step=0.2,gray!20,thin] (4,0) grid (7.0,3.0);
    \node[align=center] at (1.5,-0.3) {${f}_I$};
    \node[align=center] at (5.5,-0.3) {${f}_J$};
					
	\draw[-,thick] (0,0) -- (0,3.0);
	\draw[-,thick] (0,0) -- (3.0,0);
	\draw[-,thick] (3.0,0) -- (3.0,3.0);
	\draw[-,thick] (0,3.0) -- (3.0,3.0);
	
	\draw[-,thick] (4,0) -- (4,3);
	\draw[-,thick] (4,0) -- (7,0);
	\draw[-,thick] (7,0) -- (7,3);
	\draw[-,thick] (4,3) -- (7,3);
	
	\filldraw[fill=blue!40!white, draw=black] (1.4,1.6) rectangle (1.6,1.4);	
	
	\filldraw[fill=blue!40!white, draw=black] 
	(5.4,1.6) rectangle (5.6,1.4);
	\filldraw[fill=blue!40!white, draw=black] 
	(5.2,1.6) rectangle (5.4,1.4);	
	\filldraw[fill=blue!40!white, draw=black] 
	(5.6,1.6) rectangle (5.8,1.4);	
	\filldraw[fill=blue!40!white, draw=black] 
	(5.4,1.8) rectangle (5.6,1.6);
	\filldraw[fill=blue!40!white, draw=black] 
	(5.2,1.8) rectangle (5.4,1.6);	
	\filldraw[fill=blue!40!white, draw=black] 
	(5.6,1.8) rectangle (5.8,1.6);		
	\filldraw[fill=blue!40!white, draw=black] 
	(5.4,1.4) rectangle (5.6,1.2);
	\filldraw[fill=blue!40!white, draw=black] 
	(5.2,1.4) rectangle (5.4,1.2);	
	\filldraw[fill=blue!40!white, draw=black] 
	(5.6,1.4) rectangle (5.8,1.2);		
	
	\filldraw[fill=blue!40!white, draw=black] 
	(5.4,2.0) rectangle (5.6,1.8);	
	\filldraw[fill=blue!40!white, draw=black] 
	(5.0,2.0) rectangle (5.2,1.8);		
	\filldraw[fill=blue!40!white, draw=black] 
	(4.6,2.0) rectangle (4.8,1.8);
	\filldraw[fill=blue!40!white, draw=black] 
	(5.8,2.0) rectangle (6.0,1.8);		
	\filldraw[fill=blue!40!white, draw=black] 
	(6.2,2.0) rectangle (6.4,1.8);	
	
	\filldraw[fill=blue!40!white, draw=black] 
	(5.4,2.4) rectangle (5.6,2.2);	
	\filldraw[fill=blue!40!white, draw=black] 
	(5.0,2.4) rectangle (5.2,2.2);		
	\filldraw[fill=blue!40!white, draw=black] 
	(4.6,2.4) rectangle (4.8,2.2);
	\filldraw[fill=blue!40!white, draw=black] 
	(5.8,2.4) rectangle (6.0,2.2);		
	\filldraw[fill=blue!40!white, draw=black] 
	(6.2,2.4) rectangle (6.4,2.2);	

	\filldraw[fill=blue!40!white, draw=black] 
	(5.4,1.6) rectangle (5.6,1.4);	
	\filldraw[fill=blue!40!white, draw=black] 
	(5.0,1.6) rectangle (5.2,1.4);		
	\filldraw[fill=blue!40!white, draw=black] 
	(4.6,1.6) rectangle (4.8,1.4);
	\filldraw[fill=blue!40!white, draw=black] 
	(5.8,1.6) rectangle (6.0,1.4);		
	\filldraw[fill=blue!40!white, draw=black] 
	(6.2,1.6) rectangle (6.4,1.4);	
	
	\filldraw[fill=blue!40!white, draw=black] 
	(5.4,1.2) rectangle (5.6,1.0);	
	\filldraw[fill=blue!40!white, draw=black] 
	(5.0,1.2) rectangle (5.2,1.0);		
	\filldraw[fill=blue!40!white, draw=black] 
	(4.6,1.2) rectangle (4.8,1.0);
	\filldraw[fill=blue!40!white, draw=black] 
	(5.8,1.2) rectangle (6.0,1.0);		
	\filldraw[fill=blue!40!white, draw=black] 
	(6.2,1.2) rectangle (6.4,1.0);	
	
	\filldraw[fill=blue!40!white, draw=black] 
	(5.4,0.8) rectangle (5.6,0.6);	
	\filldraw[fill=blue!40!white, draw=black] 
	(5.0,0.8) rectangle (5.2,0.6);		
	\filldraw[fill=blue!40!white, draw=black] 
	(4.6,0.8) rectangle (4.8,0.6);
	\filldraw[fill=blue!40!white, draw=black] 
	(5.8,0.8) rectangle (6.0,0.6);		
	\filldraw[fill=blue!40!white, draw=black] 
	(6.2,0.8) rectangle (6.4,0.6);		
	
	\filldraw[fill=blue!40!white, draw=black] 
	(5.4,0.2) rectangle (5.6,0.0);
	\filldraw[fill=blue!40!white, draw=black] 
	(4.0,0.2) rectangle (4.2,0.0);
	\filldraw[fill=blue!40!white, draw=black] 
	(6.8,0.2) rectangle (7.0,0.0);	

	\filldraw[fill=blue!40!white, draw=black] 
	(5.4,1.6) rectangle (5.6,1.4);
	\filldraw[fill=blue!40!white, draw=black] 
	(4.0,1.6) rectangle (4.2,1.4);
	\filldraw[fill=blue!40!white, draw=black] 
	(6.8,1.6) rectangle (7.0,1.4);		

	\filldraw[fill=blue!40!white, draw=black] 
	(5.4,3.0) rectangle (5.6,2.8);
	\filldraw[fill=blue!40!white, draw=black] 
	(4.0,3.0) rectangle (4.2,2.8);
	\filldraw[fill=blue!40!white, draw=black] 
	(6.8,3.0) rectangle (7.0,2.8);	
	\end{scope}
	\end{tikzpicture}

%% file: img/g.tex
	\begin{tikzpicture}[auto,node distance=0.35cm,thick]

	\node [module,minimum width=3.2cm] (c1) at (0,0) 
	{\scriptsize{Conv $128\times 3\times 3$ , stride 1}}; 
	\node [module,minimum width=3.2cm,below=of c1] (c2)  
	{\scriptsize{Conv $192\times 3\times 3$, stride 2}}; 
	\node [module,minimum width=3.2cm,below=of c2] (c3)  
	{\scriptsize{Conv $256\times 3\times 3$, stride 2}}; 
	\node [module,minimum width=3.2cm,below=of c3] (c4)  
	{\scriptsize{Conv $320\times 3\times 3$, stride 2}}; 
	\node [module,minimum width=3.2cm,below=of c4] (c5)  
	{\scriptsize{Conv $512\times 3\times 3$, stride 2}}; 	
	\node [module,minimum width=3.2cm,below=of c5] (c6)
	{\scriptsize{Conv $512\times 3\times 3$, stride 1}}; 	
	\node [module,minimum width=3.2cm,below=of c6] (c7)  	
	{\scriptsize{Conv $320\times 3\times 3$, stride 1}}; 	
	\node [module,minimum width=3.2cm,below=of c7] (c8)
	{\scriptsize{Conv $256\times 3\times 3$, stride 1}};
	\node [module,minimum width=3.2cm,below=of c8] (c9)
	{\scriptsize{Conv $196\times 3\times 3$, stride 1}}; 		
	\node [module,minimum width=3.2cm,below=of c9] (c10)  	
	{\scriptsize{Conv $128\times 3\times 3$, stride 1}}; 		
	\node [module,minimum width=3.2cm,below=of c10] (c11)  
	{\scriptsize{Conv $64\times 3\times 3$, stride 1}};
	\node [module,minimum width=3.2cm,below=of c11] (c12)  	
	{\scriptsize{Conv $2\times 3\times 3$, stride 1}}; 		
	\node [above=of c1] (c0)
	{\scriptsize{Relation Features}}; 		
	
	\node[left=of c1,xshift=-1.5cm] (l)  {};

	\node[below=of c12] (output) {\scriptsize{Flow}};	        	
	
	\draw[-latex] (c0) -- (c1);	
	\draw[-latex] (c1) -- (c2);	
	\draw[-latex] (c2) -- (c3);					    	
	\draw[-latex] (c3) -- (c4);					    			
	\draw[-latex] (c4) -- (c5);		
    \draw[-latex] (c5) -- (c6);
    \draw[-latex] (c6) edge node[right] {\scriptsize{Upsample $\times 2$}} (c7);
	\draw[-latex] (c7) edge node[right] {\scriptsize{Upsample $\times 2$}} (c8);	
	\draw[-latex] (c8) edge node[right] {\scriptsize{Upsample $\times 2$}} (c9);	
	\draw[-latex] (c9) edge node[right] {\scriptsize{Upsample $\times 2$}} (c10);
	\draw[-latex] (c10) -- (c11);
	\draw[-latex] (c11) -- (c12);
	\draw[-latex] (c12) -- (output);

	\draw [-latex] (c5.east) to [bend right=-90] (c6.east);
	\draw [-latex] (c4.east) to [bend right=-90] (c7.east);
	\draw [-latex] (c3.east) to [bend right=-90] (c8.east);
	\draw [-latex] (c2.east) to [bend right=-90] (c9.east);
	\draw [-latex,postaction={decorate,decoration={raise=0.1cm,text along path,text align=center,text={|\scriptsize|Residual Connection}}}] (c1.east) to [bend right=-90]  (c10.east);	
	
	\end{tikzpicture}

%% file: devon.bbl
\begin{thebibliography}{10}\itemsep=-1pt

\bibitem{bailer2015iccvflowfields}
C.~Bailer, B.~Taetz, and D.~Stricker.
\newblock Flow fields: Dense correspondence fields for highly accurate large
  displacement optical flow estimation.
\newblock {\em ICCV}, 2015.

\bibitem{beauchemin1995computation}
S.~S. Beauchemin and J.~L. Barron.
\newblock The computation of optical flow.
\newblock {\em ACM computing surveys}, 1995.

\bibitem{brox2009large}
T.~Brox, C.~Bregler, and J.~Malik.
\newblock Large displacement optical flow.
\newblock {\em CVPR}, 2009.

\bibitem{brox2004high}
T.~Brox, A.~Bruhn, N.~Papenberg, and J.~Weickert.
\newblock High accuracy optical flow estimation based on a theory for warping.
\newblock {\em ECCV}, 2004.

\bibitem{butler2012naturalistic}
D.~J. Butler, J.~Wulff, G.~B. Stanley, and M.~J. Black.
\newblock A naturalistic open source movie for optical flow evaluation.
\newblock {\em ECCV}, 2012.

\bibitem{chen2016deeplab}
L.-C. Chen, G.~Papandreou, I.~Kokkinos, K.~Murphy, and A.~L. Yuille.
\newblock Deeplab: Semantic image segmentation with deep convolutional nets,
  atrous convolution, and fully connected crfs.
\newblock {\em arXiv}, 2016.

\bibitem{dai17dcn}
J.~Dai, H.~Qi, Y.~Xiong, Y.~Li, G.~Zhang, H.~Hu, and Y.~Wei.
\newblock Deformable convolutional networks.
\newblock {\em ICCV}, 2017.

\bibitem{fischer2015flownet}
A.~Dosovitskiy, P.~Fischer, E.~Ilg, P.~H{\"a}usser, C.~Haz{\i}rba{\c{s}},
  V.~Golkov, P.~van~der Smagt, D.~Cremers, and T.~Brox.
\newblock Flownet: Learning optical flow with convolutional networks.
\newblock {\em ICCV}, 2015.

\bibitem{fleet2006optical}
D.~Fleet and Y.~Weiss.
\newblock Optical flow estimation.
\newblock In {\em Handbook of mathematical models in computer vision}. 2006.

\bibitem{fortun2015optical}
D.~Fortun, P.~Bouthemy, and C.~Kervrann.
\newblock Optical flow modeling and computation: a survey.
\newblock {\em CVIU}, 2015.

\bibitem{Geiger2012CVPR}
A.~Geiger, P.~Lenz, and R.~Urtasun.
\newblock Are we ready for autonomous driving? the kitti vision benchmark
  suite.
\newblock {\em CVPR}, 2012.

\bibitem{he2015delving}
K.~He, X.~Zhang, S.~Ren, and J.~Sun.
\newblock Delving deep into rectifiers: Surpassing human-level performance on
  imagenet classification.
\newblock {\em CVPR}, 2015.

\bibitem{he2016deep}
K.~He, X.~Zhang, S.~Ren, and J.~Sun.
\newblock Deep residual learning for image recognition.
\newblock {\em CVPR}, 2016.

\bibitem{horn1981determining}
B.~K. Horn and B.~G. Schunck.
\newblock Determining optical flow.
\newblock {\em Artificial Intelligence}, 1981.

\bibitem{hui2018liteflownet}
T.-W. Hui, X.~Tang, and C.~C. Loy.
\newblock Liteflownet: A lightweight convolutional neural network for optical
  flow estimation.
\newblock {\em CVPR}, 2018.

\bibitem{hur2015generalized}
J.~Hur, H.~Lim, C.~Park, and S.~Chul~Ahn.
\newblock Generalized deformable spatial pyramid: Geometry-preserving dense
  correspondence estimation.
\newblock {\em CVPR}, 2015.

\bibitem{ilg2016flownet}
E.~Ilg, N.~Mayer, T.~Saikia, M.~Keuper, A.~Dosovitskiy, and T.~Brox.
\newblock Flownet 2.0: Evolution of optical flow estimation with deep networks.
\newblock {\em CVPR}, 2017.

\bibitem{jaderberg2015spatial}
M.~Jaderberg, K.~Simonyan, A.~Zisserman, et~al.
\newblock Spatial transformer networks.
\newblock {\em NIPS}, 2015.

\bibitem{janai2018unsupervised}
J.~Janai, F.~G{\"u}ney, A.~Ranjan, M.~Black, and A.~Geiger.
\newblock Unsupervised learning of multi-frame optical flow with occlusions.
\newblock {\em ECCV}, 2018.

\bibitem{kim2013deformable}
J.~Kim, C.~Liu, F.~Sha, and K.~Grauman.
\newblock Deformable spatial pyramid matching for fast dense correspondences.
\newblock {\em CVPR}, 2013.

\bibitem{kingma2014adam}
D.~Kingma and J.~Ba.
\newblock Adam: A method for stochastic optimization.
\newblock {\em ICLR}, 2014.

\bibitem{lu2017learning}
Y.~Lu, Z.~Yang, J.~Kannala, and S.~Kaski.
\newblock Learning image relations with contrast association networks.
\newblock {\em arXiv}, 2017.

\bibitem{lucas1981iterative}
B.~D. Lucas and T.~Kanade.
\newblock An iterative image registration technique with an application to
  stereo vision.
\newblock {\em IJCAI}, 1981.

\bibitem{mayer2018makes}
N.~Mayer, E.~Ilg, P.~Fischer, C.~Hazirbas, D.~Cremers, A.~Dosovitskiy, and
  T.~Brox.
\newblock What makes good synthetic training data for learning disparity and
  optical flow estimation?
\newblock {\em IJCV}, 2018.

\bibitem{mayer2016large}
N.~Mayer, E.~Ilg, P.~Hausser, P.~Fischer, D.~Cremers, A.~Dosovitskiy, and
  T.~Brox.
\newblock A large dataset to train convolutional networks for disparity,
  optical flow, and scene flow estimation.
\newblock {\em CVPR}, 2016.

\bibitem{ranjan2016optical}
A.~Ranjan and M.~J. Black.
\newblock Optical flow estimation using a spatial pyramid network.
\newblock {\em CVPR}, 2017.

\bibitem{epicflow}
J.~Revaud, P.~Weinzaepfel, Z.~Harchaoui, and C.~Schmid.
\newblock {EpicFlow: Edge-Preserving Interpolation of Correspondences for
  Optical Flow}.
\newblock {\em CVPR}, 2015.

\bibitem{ronneberger2015u}
O.~Ronneberger, P.~Fischer, and T.~Brox.
\newblock U-net: Convolutional networks for biomedical image segmentation.
\newblock {\em MICCAI}, 2015.

\bibitem{roth2007spatial}
S.~Roth and M.~J. Black.
\newblock On the spatial statistics of optical flow.
\newblock {\em IJCV}, 2007.

\bibitem{scharstein2002taxonomy}
D.~Scharstein and R.~Szeliski.
\newblock A taxonomy and evaluation of dense two-frame stereo correspondence
  algorithms.
\newblock {\em IJCV}, 2002.

\bibitem{sevilla2014optical}
L.~Sevilla-Lara, D.~Sun, E.~G. Learned-Miller, and M.~J. Black.
\newblock Optical flow estimation with channel constancy.
\newblock {\em ECCV}, 2014.

\bibitem{sun2018models}
D.~Sun, X.~Yang, M.-Y. Liu, and J.~Kautz.
\newblock Models matter, so does training: An empirical study of cnns for
  optical flow estimation.
\newblock {\em arXiv}, 2018.

\bibitem{sun2017pwc}
D.~Sun, X.~Yang, M.-Y. Liu, and J.~Kautz.
\newblock Pwc-net: Cnns for optical flow using pyramid, warping, and cost
  volume.
\newblock {\em CVPR}, 2018.

\bibitem{teney2016learning}
D.~Teney and M.~Hebert.
\newblock Learning to extract motion from videos in convolutional neural
  networks.
\newblock {\em arXiv}, 2016.

\bibitem{thewlis2016fully}
J.~Thewlis, S.~Zheng, P.~H. Torr, and A.~Vedaldi.
\newblock Fully-trainable deep matching.
\newblock {\em BMVC}, 2016.

\bibitem{ummenhofer2016demon}
B.~Ummenhofer, H.~Zhou, J.~Uhrig, N.~Mayer, E.~Ilg, A.~Dosovitskiy, and
  T.~Brox.
\newblock Demon: Depth and motion network for learning monocular stereo.
\newblock {\em CVPR}, 2017.

\bibitem{wang2018occlusion}
Y.~Wang, Y.~Yang, Z.~Yang, L.~Zhao, and W.~Xu.
\newblock Occlusion aware unsupervised learning of optical flow.
\newblock {\em CVPR}, 2018.

\bibitem{mrflow}
J.~Wulff, L.~Sevilla-Lara, and M.~J. Black.
\newblock Optical flow in mostly rigid scenes.
\newblock {\em CVPR}, 2017.

\bibitem{XuCVPR2017DCFlow}
J.~Xu, R.~Ranftl, and V.~Koltun.
\newblock {Accurate Optical Flow via Direct Cost Volume Processing}.
\newblock {\em CVPR}, 2017.

\bibitem{yang2017s2f}
Y.~Yang and S.~Soatto.
\newblock S2f: Slow-to-fast interpolator flow.
\newblock {\em CVPR}, 2017.

\bibitem{yu2015multi}
F.~Yu and V.~Koltun.
\newblock Multi-scale context aggregation by dilated convolutions.
\newblock {\em ICLR}, 2015.

\bibitem{zbontar2016stereo}
J.~Zbontar and Y.~LeCun.
\newblock Stereo matching by training a convolutional neural network to compare
  image patches.
\newblock {\em JMLR}, 2016.

\end{thebibliography}
